\documentclass{article}

\PassOptionsToPackage{numbers, compress}{natbib}
\usepackage[preprint]{neurips_2026}

\usepackage[utf8]{inputenc} 
\usepackage[T1]{fontenc}    
\usepackage{url}            
\usepackage{booktabs}       
\usepackage{amsfonts}       
\usepackage{nicefrac}       
\usepackage{microtype}      
\usepackage{xcolor}         
\usepackage{adjustbox}
\definecolor{citecolor}{HTML}{0071BC}
\definecolor{linkcolor}{HTML}{ED1C24}
\usepackage[pagebackref=true, breaklinks=true, colorlinks,
            citecolor=citecolor, linkcolor=linkcolor, bookmarks=false]{hyperref}
\usepackage[symbol]{footmisc}
\newcommand{\specialfootnote}[1]{\footnote{\leavevmode#1}}

\usepackage{placeins}       
\usepackage{etoc}           

\etocsettagdepth{main}{none}
\etocsettagdepth{appendix}{subsection}

\usepackage{amsmath}
\usepackage{bbm}

\usepackage{graphicx} 
\usepackage{wrapfig}        
\usepackage{multirow}
\usepackage{multicol}
\usepackage{xspace}
\usepackage{arydshln}
\usepackage{enumitem}
\DeclareRobustCommand{\etal}{\emph{et~al.}\@\xspace}

\newcommand{\hiddensection}[1]{%
  \refstepcounter{section}%
  \addcontentsline{toc}{section}{\protect\numberline{\thesection}#1}%
  \sectionmark{#1}%
}

\newcommand{\greencheck}{%
  \textcolor{green!60!black}{\scalebox{1.25}{\boldmath$\checkmark$}}%
}
\title{GPIC: A Giant Permissive Image Corpus \\ for Visual Generation}

\author{
Keshigeyan Chandrasegaran $^{*1}$\quad
Kyle Sargent $^{*1}$\quad
Suchir Agarwal $^{1}$\quad
Michael Jang $^{1}$\\
[0.04cm]
\textbf{Michael Poli $^{1,2}$}\quad
\textbf{Juan Carlos Niebles $^{1,4}$}\quad
\textbf{Justin Johnson $^{3}$}\quad 
\textbf{Jiajun Wu $^{1}$}\quad 
\textbf{Li Fei-Fei $^{1}$} \\
[0.08cm]
$^1$~Stanford University \quad
$^2$~Radical Numerics\quad
$^3$~University of Michigan\quad
$^4$~Salesforce Research\\
[0.1cm]
\href{https://gpic.stanford.edu}{\texttt{gpic.stanford.edu}}\\
}

\begin{document}

\etocdepthtag.toc{main}

\maketitle

\begin{abstract}
    {
    Studying scalable methods for visual generative modeling requires large, accessible, and stable datasets.
    We introduce \textbf{GPIC}, a \textbf{G}iant \textbf{P}ermissive \textbf{I}mage \textbf{C}orpus of approximately \textbf{28 trillion pixels}.
    GPIC comprises diverse internet images captioned by a state-of-the-art vision-language model, including 100M training, 200K validation, and 1M test examples.
    Moreover, all GPIC images are permissively licensed for both research and commercial use.
    GPIC is safety-filtered, deduplicated, and centrally hosted on Hugging Face.
    We provide a benchmarking protocol for generative modeling on GPIC.
    Finally, we provide a reference baseline for pixel-space flow matching on GPIC.
    Our dataset, benchmark, and models
    are available on \href{https://huggingface.co/datasets/stanford-vision-lab/gpic}{Hugging Face}.}
    Evaluation toolkit and code are available at \href{https://gpic.stanford.edu}{gpic.stanford.edu}.
\end{abstract}

\footnotetext{$^*$ Equal contribution.}
\vspace{0.21ex}
\footnotetext{
~~~Correspondence to \texttt{\{keshik,ksarge\}@cs.stanford.edu}}

\begin{figure*}[p]
  \thispagestyle{empty}
  \centering
  \includegraphics[width=\textwidth]{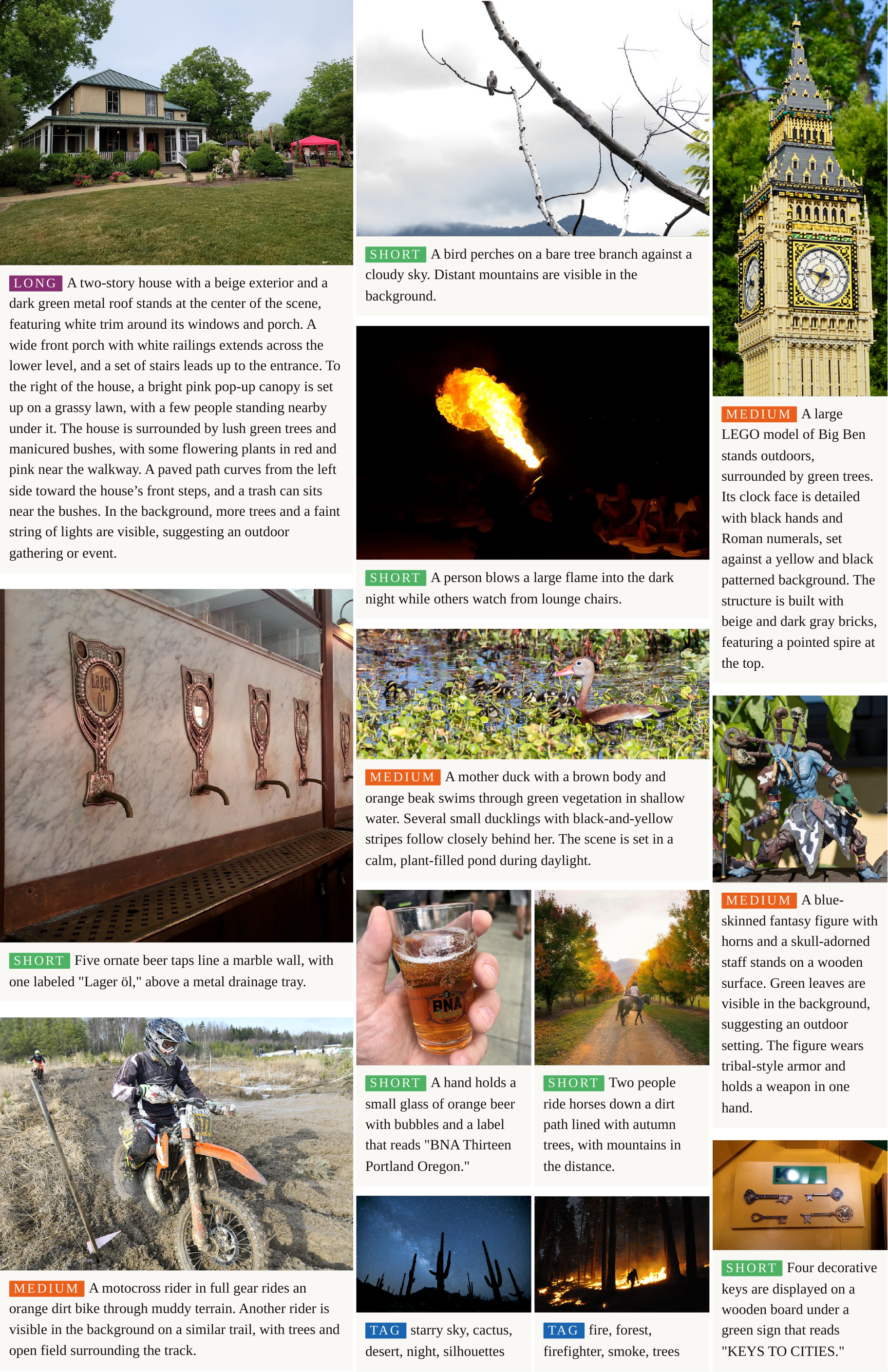}
  \caption{Example image-caption pairs from GPIC. Additional samples are shown in Figure \ref{fig:gpic_samples}.}
  \label{fig:teaser}

\end{figure*}

\section{Introduction} 

As the capabilities of modern generative models for images and videos have rapidly advanced, so has their appetite for data. Although the training details of frontier visual generative models are seldom made public, state-of-the-art open-weight models are trained on image and video corpora containing hundreds of millions to billions of examples~\cite{wan2025wan,rombach2022high,saharia2022photorealistic,zhang2026qwen}. Proprietary models presumably operate at comparable or greater data scales~\cite{videoworldsimulators2024,google2026nanobanana2}. In addition, visual generation has shifted away from class-conditioning signals toward dense conditioning signals such as rich text captions~\cite{saharia2022photorealistic,videoworldsimulators2024}.

\renewcommand{\thefootnote}{\arabic{footnote}} 
The class-conditional ImageNet-1K benchmark has driven substantial progress in visual generation research, serving as a testbed for methods such as BigGAN~\cite{biggan}, VQVAE~\cite{vqvae}, VQGAN~\cite{vqgan}, and DiT~\cite{dit}. 
However, after more than a decade of focus on the same visual generation benchmark, two critical issues have become apparent.
First, modern visual generative models rely on much larger and more diverse training corpora together with rich conditioning signals such as free-form text. As a result, the ImageNet-1K benchmark has increasingly drifted away from contemporary visual generative modeling practice, with conclusions less likely to transfer to modern practical settings. 
Second, over a decade of hillclimbing on ImageNet-1K has saturated FID scores and driven ``Goodharting'' of the metric{\setlength{\footnotemargin}{2.2em}
\specialfootnote{~Goodhart's law: \textit{When a measure becomes a target, it ceases to be a good measure.}}}.
Notably, several recent methods achieve lower FID scores on the ImageNet-1K benchmark than held-out real images~\cite{rae, simpler_diffusion, lightningdit, var}.

On the other hand, many industrial labs report results on text-conditioned generation of images and video, but use proprietary or unstable datasets, hindering reproducibility and open scientific comparisons. This motivates rethinking benchmark datasets for visual generative modeling research. Concretely, we identify four key properties of a modern benchmark dataset for visual generation.

\begin{itemize}[leftmargin=18pt]
    \item \textbf{Permissive:} Every image in the dataset should have a known license permitting both research and commercial use, without imposing restrictions on derived artifacts. Moreover, the dataset itself, including metadata and annotations, should be released under a permissive license.
    
    \item \textbf{Stable:} To ensure valid scientific comparisons, the benchmark dataset cannot change over time. Many modern image datasets are distributed as URL indices, which makes comparisons difficult due to link rot~\cite{datacomp, yfcc100m}.

    \item \textbf{Large:} The benchmark dataset should be large enough, with rich text captions, to train and evaluate modern visual generative models.

    \item \textbf{Accessible:} The dataset must be easily downloadable in a sharded format without requiring crawling infrastructure~\cite{laion, datacomp, img2dataset} or memory-intensive resharding~\cite{yfcc100m}.
\end{itemize}

We introduce \textbf{GPIC}, a \textbf{G}iant \textbf{P}ermissive \textbf{I}mage \textbf{C}orpus designed to satisfy all four criteria for benchmarking visual generative models~(Table \ref{tab:dataset-properties}). GPIC comprises 27.97 trillion pixels across 100M training, 200K validation, and 1M test examples captioned with Qwen3-VL-4B~\cite{bai2025qwen3vl}.
GPIC is centrally hosted on Hugging Face as 8,000 shards, providing stable and accessible infrastructure for large-scale training. To construct GPIC, we develop pipelines for licensed image crawling, large-scale captioning, safety and quality filtering, and deduplication (Section ~\ref{sec:construction}). 
We also revisit the ImageNet-1K evaluation protocol~(Figure \ref{fig:fid-saturation}), providing a new benchmarking protocol based on FD-DINOv2~\cite{stein2023exposing} against a held-out set of one million GPIC images. 
Finally, we provide a reference pixel-space flow matching baseline on GPIC~(Section ~\ref{sec:baselines}).
We hope GPIC enables open, accessible, and reproducible research in visual generative modeling.

\begin{table}[t]
\footnotesize
\centering
\begin{adjustbox}{width=0.99\linewidth}
\begin{tabular}{lccccc}
\toprule
\textbf{Property} & {ImageNet-1K} \cite{imagenet} & {YFCC100M} \cite{yfcc100m} & {OpenImages} \cite{openimages} & {DataComp} \cite{datacomp} & \textbf{GPIC} \\
\midrule
\textbf{Permissive} & & & \greencheck & & \greencheck \\
\midrule
\textbf{Stable}     & \greencheck & \greencheck & & & \greencheck \\
\midrule
\textbf{Large}      & & \greencheck & & \greencheck & \greencheck \\
\midrule
\textbf{Accessible} & \greencheck & & \greencheck & & \greencheck \\
\bottomrule
\end{tabular}
\end{adjustbox}
\vspace{2mm}
\caption{
Existing image benchmark datasets fail to satisfy all four criteria. \textbf{GPIC} satisfies all four criteria.
}
\vspace{-0.5cm}
\label{tab:dataset-properties}
\end{table}

\begin{figure}[h]
\begin{adjustbox}{width=1\textwidth,center}
\begin{tabular}{c}
    \includegraphics[width=1\textwidth]
    {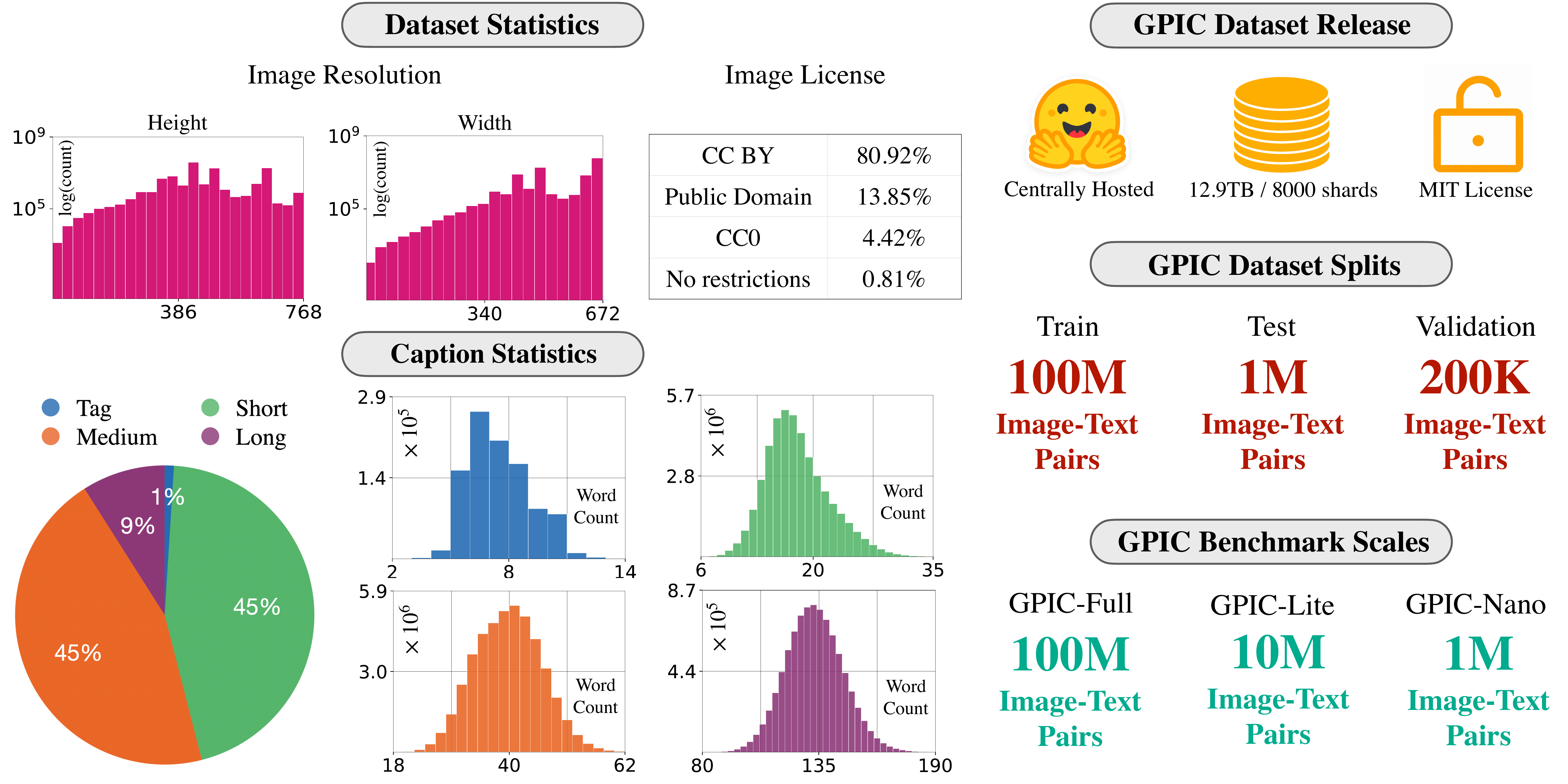} \\
\end{tabular}
\end{adjustbox}
\caption{
\textbf{GPIC dataset statistics.}
The figure shows GPIC's image height and width distributions, license composition, caption statistics, release format, dataset splits, and benchmark scales.
GPIC images have an average height of 479 pixels and an average width of 587 pixels.
GPIC is centrally hosted on Hugging Face as 8,000 shards totaling 12.9TB and released under the MIT license.
GPIC-Lite (10M) and GPIC-Nano (1M) provide smaller subsets for development.
Best viewed in color.
}
\label{fig_main:samples}
\vspace{-0.2cm}
\end{figure}

\section{Dataset Construction}
\label{sec:construction}

\begin{figure}[h]
\centering
\includegraphics[width=0.99\textwidth]{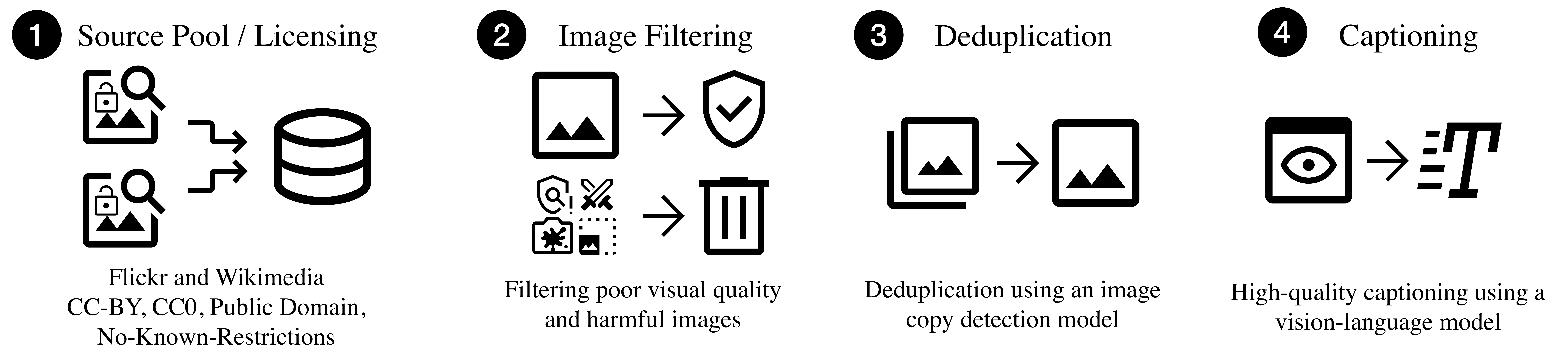}
\caption{\textbf{Our dataset construction pipeline.}
We develop a four-stage pipeline to create GPIC. We source permissive images from Flickr and Wikimedia (Stage 1), filter low-quality and harmful images (Stage 2), deduplicate images using similarity scores derived from SSCD~\cite{pizzi2022self} copy detection features (Stage 3), and caption into one of tag, short, medium, or long (Stage 4). Qwen-3-VL-4B-Instruct~\cite{bai2025qwen3vl} is used for filtering and captioning.
}
\label{fig:data_pipeline}
\end{figure}

In this section, we provide an overview of the GPIC construction pipeline (Figure~\ref{fig:data_pipeline}).

\subsection{Source Pool and Licensing}
We construct GPIC by collecting images under permissive licenses that allow redistribution and commercial use. We source images from Flickr and Wikimedia, restricting the source pool to CC BY, CC0, Public Domain, and No-Known-Restrictions categories. This licensing criterion ensures that GPIC can be used by both academic and industrial researchers without restricting the release or downstream use of derived artifacts. For each retrieved image, we retain provenance and attribution metadata, including a dataset-generated key, image height and width, retrieval timestamp, license name, license URL, and attribution string. The final dataset excludes retrieved image URLs, avoiding release of a large-scale URL index while preserving attribution and license information. The initial source pool contains 110,569,761 images, with 87.7\% sourced from Flickr and 12.3\% from Wikimedia.

\subsection{Image Filtering}
\label{sec:construction:filtering}
\begin{wrapfigure}[13]{l}[0pt]{0.32\textwidth}
\vspace{-0.45cm}
\begin{adjustbox}{width=0.95\linewidth,center}
\begin{tabular}{c}
    \includegraphics[width=0.99\textwidth]{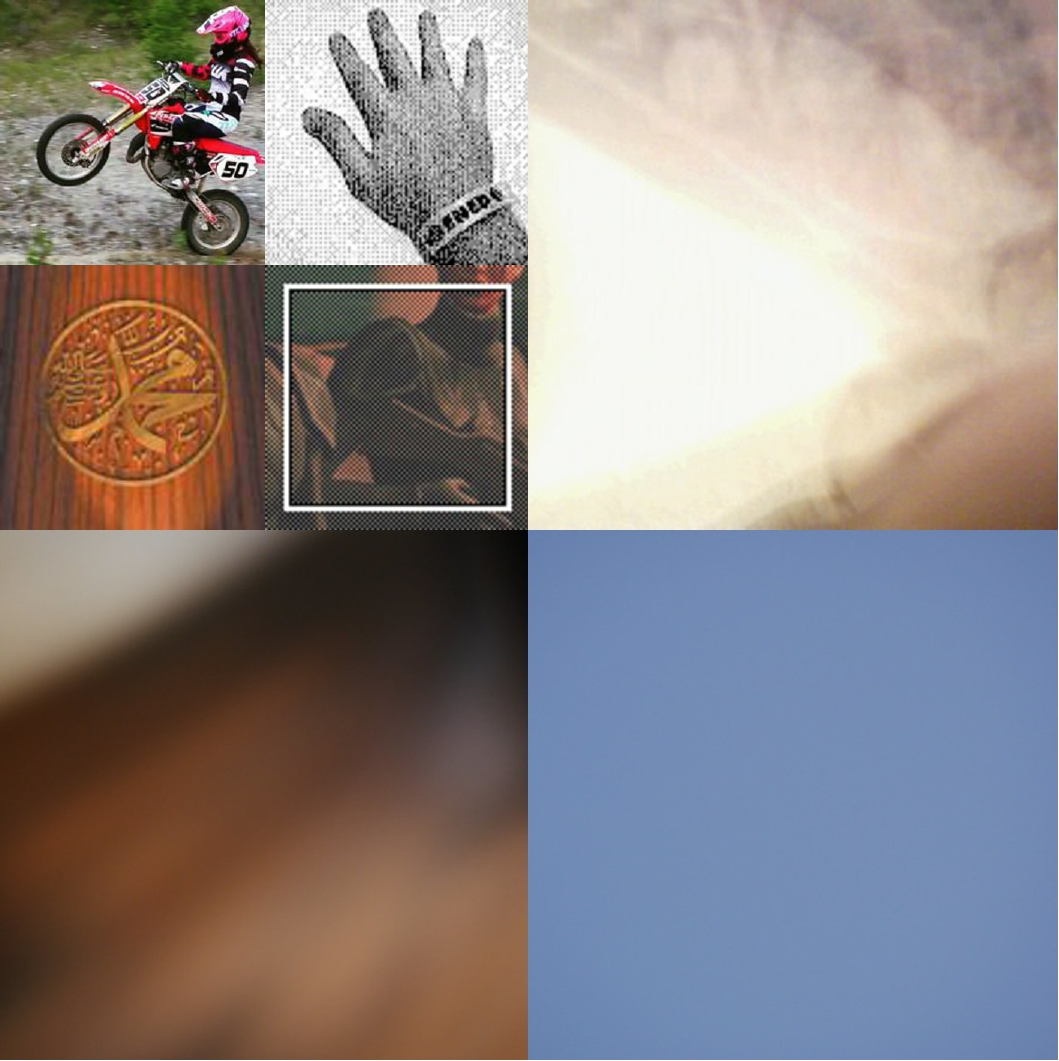} \\
\end{tabular}
\end{adjustbox}
\vspace{-0.3cm}
\caption{
Example images that are filtered due to low resolution and poor visual quality.
}
\label{fig:filtered}
\end{wrapfigure}

We apply a sequence of image-level filters to remove images unsuitable for training or benchmarking.
First, we remove images with extreme resolutions or aspect ratios.
Together, these filters remove approximately $0.01\%$ of the source pool.
We also discard images whose longest side is smaller than $256$ pixels.
Next, we apply VLM-based quality filtering using Qwen3-VL-4B.
This filter removes images with poor visual quality or limited semantic content, including near-blank images, severe blur, underexposure, and overexposure.
This stage removes approximately $0.3\%$ of the source pool. 
We show examples in Figures~\ref{fig:filtered} and \ref{fig:filtered-supp}.
Finally, we apply a conservative safety filter using Qwen3-VL-4B to remove images flagged as unsafe.
This stage removes approximately $0.35\%$ of the source pool.

\vspace{0.3cm}
\subsection{Deduplication}

\begin{figure}[t]
\begin{adjustbox}{width=1\textwidth,center}
\begin{tabular}{c}
    \includegraphics[width=1\textwidth]
    {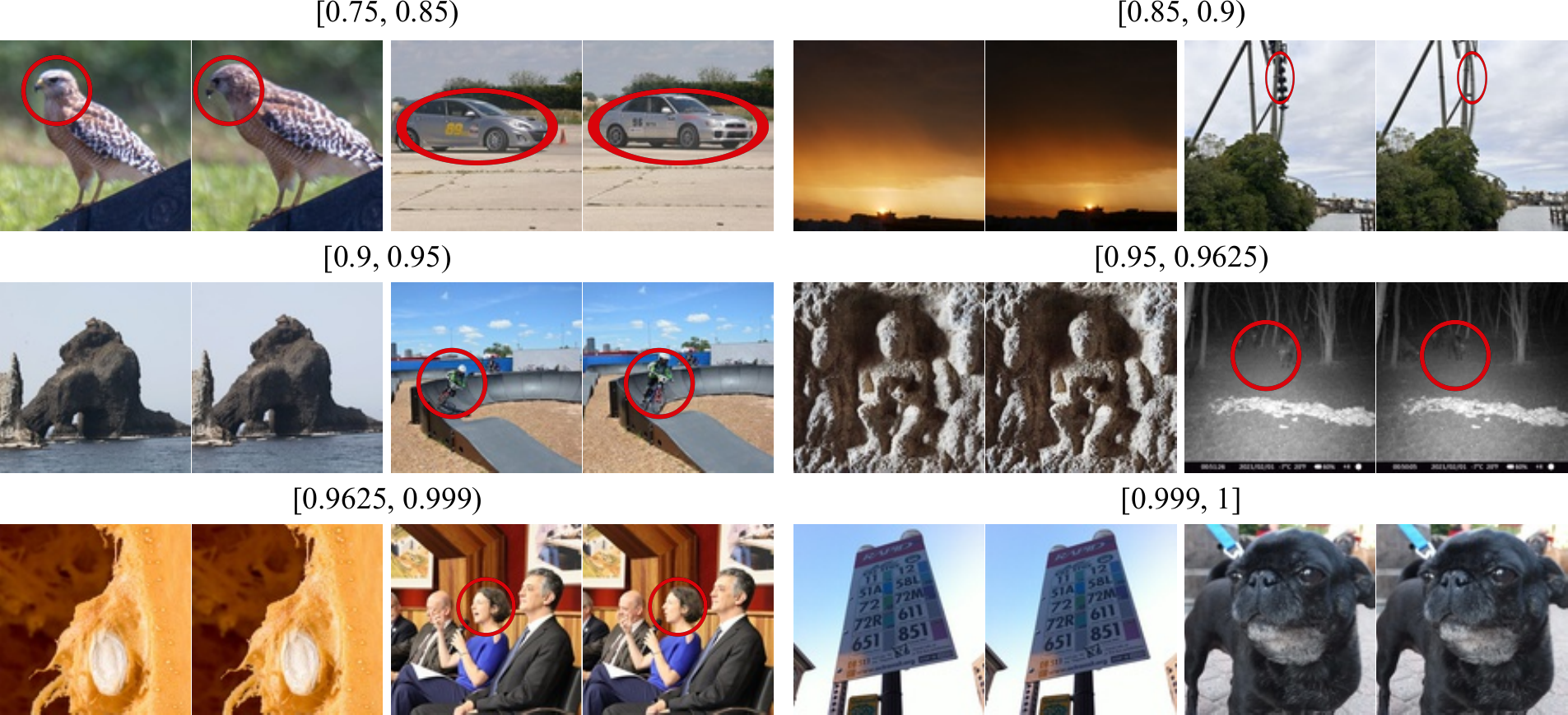} \\
\end{tabular}
\end{adjustbox}
\vspace{-0.3cm}
\caption{
\textbf{Qualitative examples of similar image pairs across SSCD similarity ranges.}
Each group shows nearest-neighbor image pairs within the indicated SSCD similarity interval.
At lower thresholds, similar pairs often contain visually related but distinct images, including changes in pose, viewpoint, or object identity.
At higher thresholds, pairs increasingly correspond to near-duplicates, but visible differences can still remain (highlighted in red).
Together with the high cost of obtaining permissively licensed images at scale, these examples motivate conservative duplicate removal rather than aggressive thresholding.
Best viewed in color.
}
\label{fig:similarity-ranges}
\vspace{-1cm}
\end{figure}

GPIC is built from Flickr and Wikimedia, where duplicated visual content naturally arises from burst photography, reposts, and edited variants of memes and viral images. Since permissively licensed images are costly to obtain at scale, we adapt conservative duplicate removal: removing clear duplicates and near-duplicates while retaining visually related but distinct images.
Many duplicates are not pixel-identical, so we perform deduplication using copy-detection features. Specifically, we extract SSCD features~\cite{pizzi2022self} for all images and use FAISS for  approximate nearest-neighbor search. 
We first manually inspect nearest-neighbor pairs across SSCD similarity ranges to calibrate removal thresholds. 
This inspection shows that similarity above 0.90 often indicates substantial shared visual content, but still includes distinct images with changes in pose, viewpoint, or scene composition. Even pairs between 0.95 and 0.9625 can remain visually distinct, so we avoid removing images from individual pairs unless their similarity exceeds a more conservative threshold (Figure~\ref{fig:similarity-ranges}).

\textbf{Image collision models.}
Full-corpus deduplication at 110M-image scale is expensive, and threshold\begin{wrapfigure}[22]{r}[0pt]{0.47\textwidth}
\vspace{-0.3cm}
\begin{adjustbox}{width=0.99\linewidth,center}
\begin{tabular}{c}
    \includegraphics[width=0.99\textwidth]{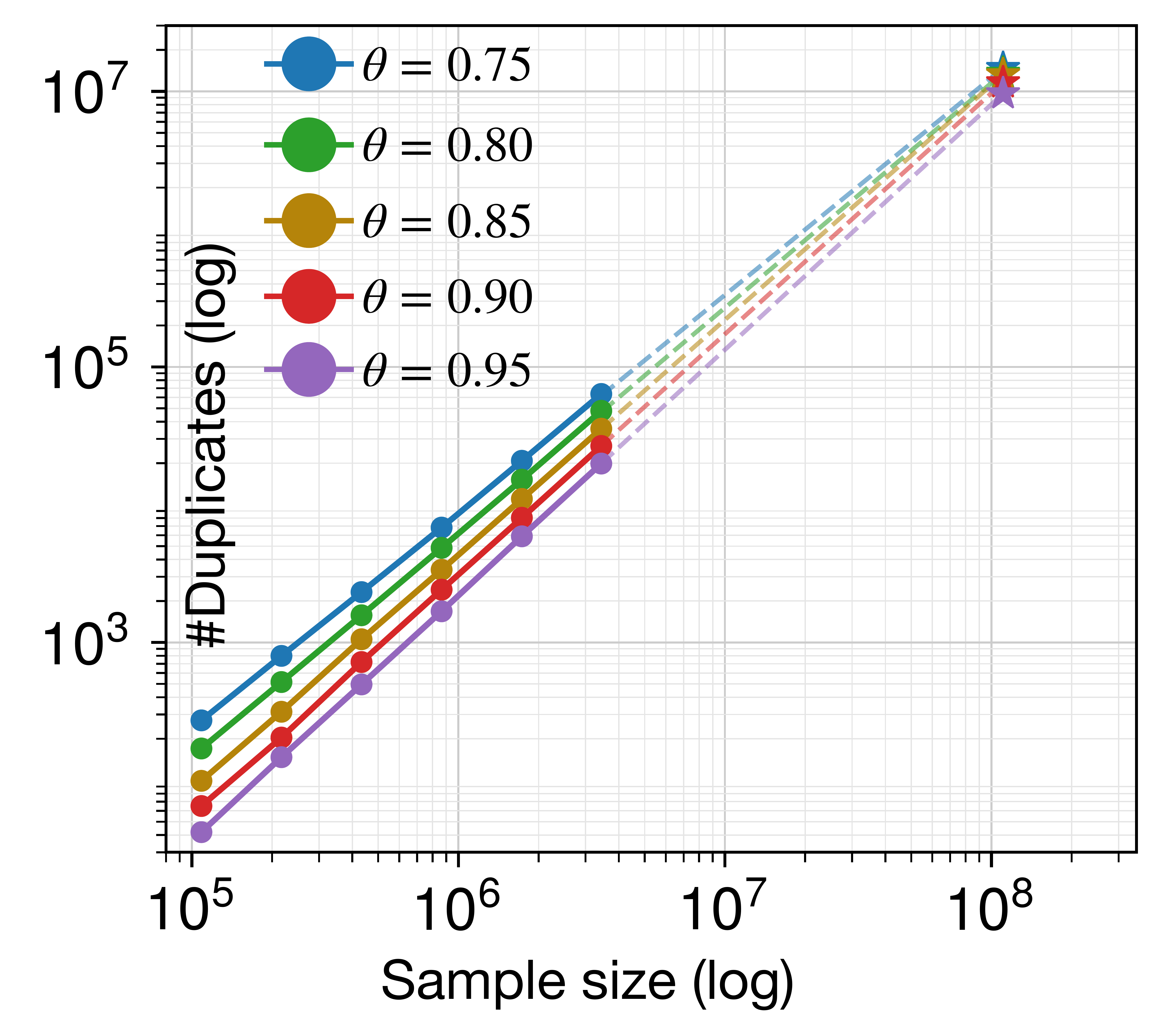} \\
\end{tabular}
\end{adjustbox}
\vspace{-0.3cm}
\caption{
\textbf{Image collision models.}
SSCD-based duplicate removals follow a power-law trend across subset sizes and similarity thresholds. Extrapolating to the 110M-image source pool shows 
that
\(\theta=0.95\) is estimated to remove \(9.62\times10^6\) images, leaving approximately \(1.01\times10^8\) images.
}
\label{fig_main:collision_models}
\end{wrapfigure}choice strongly affects how many images are removed.
We therefore build predictive collision models on smaller subsets before running the final full-corpus pass.
We run SSCD-based deduplication on six subsets ranging from 108K to 3.4M images, across thresholds \(\theta\in\{0.75,0.80,0.85,0.90,0.95\}\).
For each threshold, we connect nearest-neighbor pairs whose SSCD cosine similarity exceeds \(\theta\), count the number of images that would be removed by retaining the highest-resolution image in each connected component, and fit a power law \(D(N)=AN^\beta\) to predict removals at full scale.
The resulting curves are shown in Figure~\ref{fig_main:collision_models}.
These models show that lower thresholds would remove too many images at full scale, while \(\theta=0.95\) provides a conservative operating point.
At \(\theta=0.95\), the model estimates \(9.62\times10^6\) removed images, leaving approximately \(1.01\times10^8\) images for the final release pipeline.

\textbf{Full-corpus deduplication.}
Rather than applying a single threshold of \(0.95\) to remove images from all similar pairs, we use a more conservative two-tier rule calibrated by manual inspection.
We first construct a candidate similarity graph by connecting image pairs with SSCD similarity above \(0.90\).
Within this graph, we apply two removal rules.
First, for pairs with similarity above \(0.9625\), we remove the lower-resolution image, targeting high-confidence duplicate pairs.
Second, for connected components containing at least five images, we keep only the highest-resolution image in the component, targeting repeated near-copy clusters.
This rule prioritizes avoiding false removals of distinct images while still removing high-confidence duplicates and large repeated clusters.
After deduplication, \(101.3\)M images remain.
We show examples in Figure~\ref{fig:dedup-examples}.
We also verify that no exact duplicates remain by computing SHA-256 hashes over image file bytes.

\subsection{Captioning GPIC}
\label{sec:construction:captioning}

GPIC uses high-quality synthetic captions generated by a vision-language model rather than source metadata or alt text, which are often unavailable, noisy, or weakly aligned with image content.

\paragraph{Caption formats.}
There are many valid ways to describe an image in words, ranging from unordered keywords to detailed scene descriptions.
To capture this variation, GPIC uses four caption formats: tag, short, medium, and long.
Tag captions are unordered keyword lists, while short, medium, and long captions provide increasingly detailed natural-language descriptions of an image.
Examples are shown in Figure~\ref{fig_main:samples}.
In the final corpus, caption types are assigned with proportions 1\% tag, 45\% short, 45\% medium, and 9\% long.

\begin{figure}[b]
\centering
\includegraphics[width=0.99\textwidth]{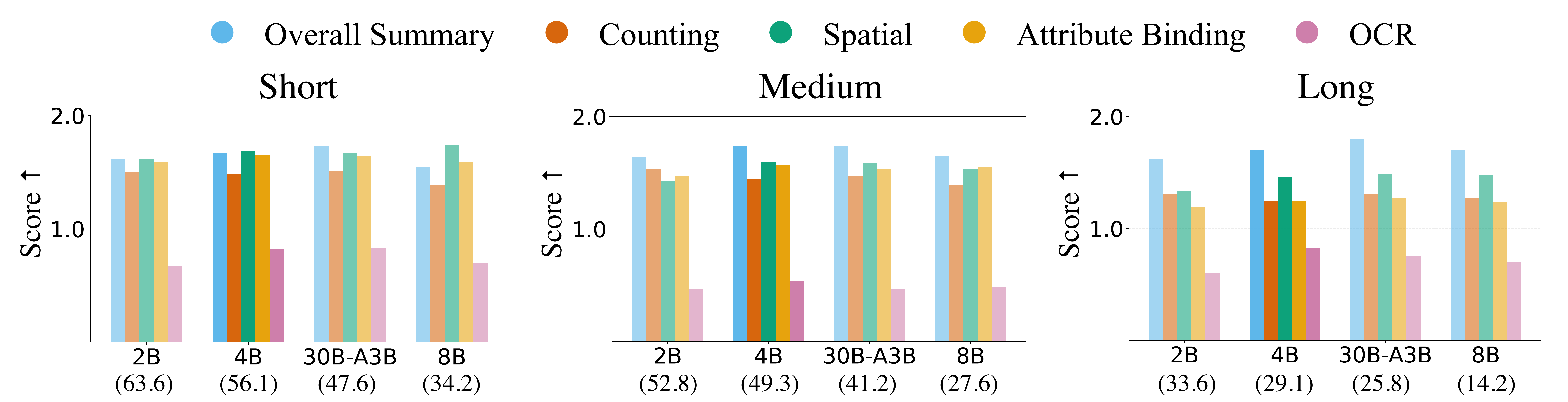}
\vspace{-0.25cm}
\caption{
\textbf{Captioning model selection.}
We evaluate Qwen3-VL-Instruct models on the GPIC captioning microbenchmark across five caption-quality criteria and throughput.
Throughput in images per second (1xH100) is shown in parentheses below each model.
Qwen3-VL-4B-Instruct provides the best quality-throughput tradeoff: it matches or approaches the best quality scores across short, medium, and long captions while maintaining higher throughput than larger models.
}
\vspace{-0.2cm}
\label{fig:microbenchmark}
\end{figure}

\textbf{Captioning model selection.}
Captioning 100M images requires a model that is accurate, fast, and practical to run at scale.
Closed-source VLMs are prohibitively expensive for full-corpus captioning, so we focus on open-source models.
We consider Qwen3-VL models~\cite{bai2025qwen3vl} because they are among the strongest open-source models for image understanding, are available at multiple scales, and support efficient inference through standard serving frameworks such as vLLM~\cite{kwon2023efficient} and SGLang~\cite{zheng2024sglang}.
Existing VLM benchmarks do not directly measure the captioning capability required for GPIC, where captions must be generated at multiple levels of detail.
We therefore construct a microbenchmark of 1,520 GPIC images, covering 720 short, 640 medium, and 160 long captions.
For each image, human annotators refine initial VLM-generated captions to produce reference captions.

We evaluate Qwen3-VL-Instruct models at 2B, 4B, 8B, and 30B-A3B (sparse MoE) on this benchmark.
All captions are generated from the full image without cropping.
We score captions along five axes: overall summary quality, counting accuracy, spatial understanding, attribute binding, and OCR.
Each axis is scored on a 0--2 scale using an LLM-as-a-judge pipeline, and we also measure captioning throughput.
As shown in Figure~\ref{fig:microbenchmark}, Qwen3-VL-4B-Instruct provides the best quality-throughput tradeoff:
$\bullet$ strong overall summary quality compared to the largest model (1.68 vs. 1.73 for 30B-A3B);
$\bullet$ best spatial understanding and attribute binding scores (1.60 and 1.55);
$\bullet$ high short- and medium-caption throughput (56.10 and 49.31 images/sec).
Since short and medium captions make up 90\% of GPIC, throughput on these caption types is critical for full-corpus captioning.
Using Qwen3-VL-4B-Instruct, captioning the full corpus required approximately 1,500 H100 GPU-hours.

\textbf{Prompts and microbenchmark details.}
For reproducibility, we provide the tag, short, medium, and long captioning prompts in Figures~\ref{fig:prompt-caption-tag}, \ref{fig:prompt-caption-short}, \ref{fig:prompt-caption-medium}, and \ref{fig:prompt-caption-long}, respectively.
We provide the LLM-as-a-judge prompts in Figure~\ref{fig:prompt-llm-judge} and additional microbenchmark details in Appendix~\ref{sec_supp:microbenchmark}.

\subsection{Split Construction and Release}

\paragraph{Dataset splits.}
We partition GPIC into 100M training images, 200K validation images, and 1M test images.
Each split preserves the source distribution between Flickr and Wikimedia and the global caption-type distribution of 1\% tag, 45\% short, 45\% medium, and 9\% long.
This keeps the validation and test splits compositionally aligned with the training split.

\paragraph{Benchmark scales.}
We divide the GPIC train set into three nested tiers: \textbf{GPIC-Nano} with 1M images, \textbf{GPIC-Lite} with 10M images, and \textbf{GPIC-Full} with 100M images.
Nano and Lite are intended for faster iteration and smaller-scale development.
All three tiers preserve the source and caption-type distributions of GPIC-Full.
The first 80, 800, and 8000 shards correspond to GPIC-Nano, GPIC-Lite, and GPIC-Full, respectively, so switching between tiers only requires selecting the corresponding shard range.

\begin{figure}[t]
\begin{adjustbox}{width=1\textwidth,center}
\begin{tabular}{c}
    \includegraphics[width=1\textwidth]
    {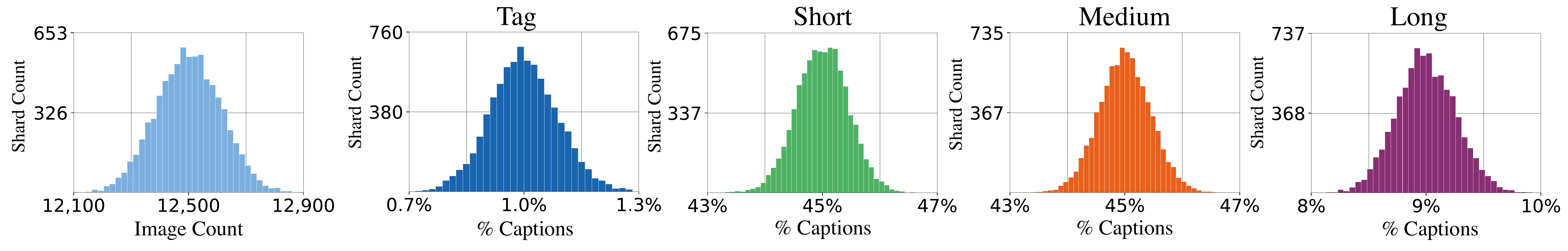} \\
\end{tabular}
\end{adjustbox}
\caption{
\textbf{GPIC shard statistics.}
We show the per-shard distribution of image counts and caption-type percentages for GPIC-Full.
GPIC-Full is shuffled into 8000 approximately  balanced shards, each containing $\approx$ 12,500 images and preserving the target caption mixture of 1\% tag, 45\% short, 45\% medium, and 9\% long captions.
\label{fig_main:per_tar}
}
\end{figure}

\paragraph{Packaging and release.}
We package GPIC as tar shards containing images, captions, and metadata, and release the dataset on Hugging Face with documentation.
To support large-scale streaming training, GPIC-Full is shuffled and organized into 8,000 balanced shards, each containing approximately 12,500 images.
As shown in Figure~\ref{fig_main:per_tar}, the shards are balanced in both image count and caption-type composition.
Each shard contains approximately 12,500 images and closely follows the target caption mixture of 1\% tag, 45\% short, 45\% medium, and 9\% long captions.
This makes each shard compositionally representative of the full corpus and avoids shard-level bias during training.

\section{Benchmarking Protocol}
Rigorous evaluation protocols are imperative to drive progress in visual generation.
A good evaluator should distinguish real and generated images while remaining aligned with human perception.
GPIC is designed to provide a more human-aligned and less saturated evaluation setting for modern visual generative models.

\begin{figure*}
    \centering
    \vspace{-3mm}
    \includegraphics[width=.94\linewidth]{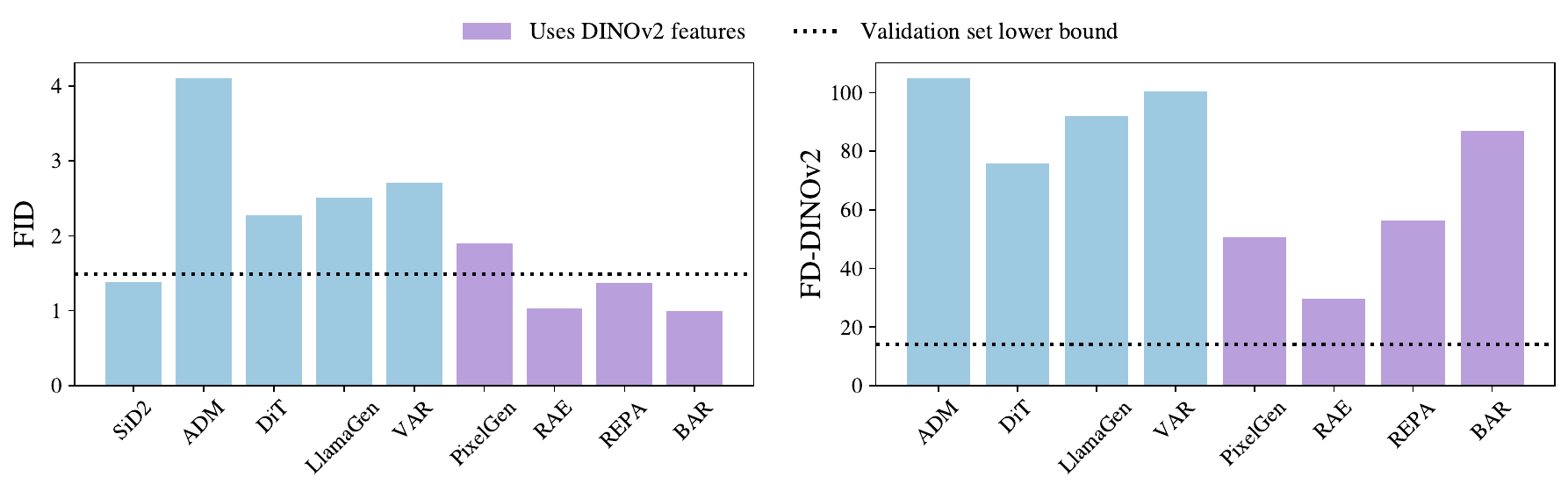}
    \vspace{-3mm}
    \caption{
    \textbf{Comparison of FID and FD-DINOv2 on ImageNet-1K.}
    ImageNet-1K FID is saturated: several models achieve lower FID than the distance between 50K held-out real ImageNet-1K images and the ImageNet-1K training set.
    By contrast, FD-DINOv2 remains unsaturated: all evaluated models have higher FD-DINOv2 than the corresponding held-out real-image distance, including models trained with DINOv2 features.
    Dotted lines indicate the distance between 50K held-out real images and the ImageNet-1K training set.
    SiD2~\cite{simpler_diffusion} is omitted from the FD-DINOv2 comparison because checkpoints or generated samples are not available.
    }
    \label{fig:fid-saturation}   
    \vspace{-4mm}
\end{figure*}

\paragraph{Metrics.}
On GPIC, we evaluate generated images using metrics computed over DINOv2 features.
Our primary metric is FD-DINOv2~\cite{stein2023exposing}, which uses the same Fr\'echet Distance formula as FID~\cite{fid} but replaces Inception features~\cite{szegedy2015going} with DINOv2 features.
We also report Precision and Density, which measure fidelity, and Recall and Coverage, which measure diversity \cite{Kynknniemi2019ImprovedPA,Sajjadi2018AssessingGM,Naeem2020ReliableFA}.
We recommend DINOv2 features because ImageNet-1K FID is saturated, while FD-DINOv2 remains informative for current models.
Figure~\ref{fig:fid-saturation} illustrates this difference.
Prior work also shows that FD-DINOv2 correlates better with human judgments than FID~\cite{stein2023exposing}.

\paragraph{Evaluation protocol.}
To evaluate a model on GPIC, users generate 50K images using the fixed set of 50K test captions that we provide, sampled randomly from the 1M GPIC test set.
The generated samples are compared against reference statistics computed from the 1M GPIC test set.
We release these precomputed test statistics on \href{https://huggingface.co/datasets/stanford-vision-lab/gpic}{Hugging Face}.
We also provide \texttt{gpic-eval}, a PyTorch evaluation suite that computes FD-DINOv2, Precision, Recall, Density, Coverage, and Maximum Mean Discrepancy as a non-parametric alternative to FD.
Additional details are provided in Appendix~\ref{sec_supp:evals}.

\paragraph{Held-out test statistics.}
GPIC also differs from standard ImageNet-1K evaluation in how reference statistics are computed.
Standard ImageNet-1K FID compares generated samples against training-set statistics.
GPIC instead compares generated samples against statistics from a held-out 1M-image test set.
This is better scientific practice because comparing against train-set statistics can fail to detect memorization or overfitting.

\paragraph{Oracle References.}
To provide reference points for interpreting generative model performance on GPIC, we report real-vs-real distances between GPIC subsets and the 1M GPIC test set.
We compute metrics for Test-50K, GPIC-Val, GPIC-Nano, GPIC-Lite, and GPIC-Full against Test-1M in Table~\ref{tab:gpic-subsets-metrics}.
These oracle references quantify the distance between real GPIC subsets under the GPIC evaluation protocol.
In particular, Test-50K vs. Test-1M provides a reference point for monitoring benchmark saturation as models trained on GPIC improve.

\textbf{GPIC Evaluation Protocol compliance.}
Use of DINOv2 features, FD-DINOv2 related loss functions, or other objectives that explicitly optimize the primary GPIC evaluation representation is \textit{strongly discouraged and must be disclosed}. Such use constitutes a material deviation from the standard GPIC protocol, since: 
\begin{itemize}
    \item DINOv2 may have been trained on images overlapping with the GPIC test set.
    \item DINOv2-based objectives directly train models to match the same representation space used by the primary GPIC metric.
\end{itemize}

Therefore, improvements in FD-DINOv2 under this setting are difficult to interpret as improvements in generative modeling capability rather than metric-specific optimization. Results that use DINOv2 or FD-DINOv2 aligned training objectives should be treated as non-standard GPIC results.

We also strongly encourage transparent reporting on the use of auxiliary networks trained on other datasets, especially larger foundation models trained on significantly more data, such as DINOv3 \cite{simeoni2025dinov3} or SigLIP \cite{zhai2023sigmoidlosslanguageimage}. Using large auxiliary models, which see considerably more data and training FLOPs, is an unfair advantage versus models trained exclusively on the GPIC benchmark dataset, and apples-to-apples comparisons are preferred whenever possible.

Other deviations from the protocol should also be reported to support interpretability and comparison across methods. These include, but are not limited to, prompt upsampling or rewriting of the provided 50K evaluation captions, and use of different captioning models or text embedding models.

\begin{table}[h]
\centering
\vspace{0.3cm}
\begin{tabular}{lccccc}
\toprule
\textbf{GPIC Subset} & \textbf{FD} $\downarrow$ & \textbf{Precision} $\uparrow$ & \textbf{Recall} $\uparrow$ & \textbf{Density} $\uparrow$ & \textbf{Coverage} $\uparrow$ \\
\midrule
Full & 1.19  & 0.947 & 0.950 & 1.000 & 0.972 \\
\midrule
Lite & 1.25 & 0.951 & 0.947 & 1.010 & 0.973 \\
\midrule
Nano & 1.60 & 0.946 & 0.946 & 1.002 & 0.968 \\
\midrule
Val  & 2.37 & 0.948 & 0.949 & 0.993 & 0.966 \\
\midrule
Test-50K & 7.44 & 0.949 & 0.953 & 0.997 & 0.967 \\
\bottomrule
\end{tabular}
\vspace{2mm}
\caption{
Oracle reference metrics over DINOv2 features for GPIC subsets evaluated against 1M GPIC Test set.
These real-vs-real values provide reference points for interpreting generative model performance on GPIC.
Metrics over Inception-v3 representations are provided in Table~\ref{tab:gpic-subsets-metrics-inception}.
Density is not upper bounded by 1.0, so values above 1.0 are valid.
}
\label{tab:gpic-subsets-metrics}
\end{table}

\clearpage
\section{Experiments}\label{sec:baselines}

We train a simple reference baseline on GPIC-Full to provide a point of comparison for future work. Our goal is not to optimize model performance, but to establish a reproducible baseline for training and evaluation on GPIC. We use JiT~\cite{li2025back}, a pixel-space flow matching model with a Transformer backbone. JiT is a natural baseline because it uses single-stage training, does not require tokenizer pretraining, and does not rely on auxiliary losses. We use the JiT-T2I (PixGen-XXL/16 1.1B) architecture proposed by Ma \etal~\cite{ma2026pixelgen}, which uses Qwen3-1.7B~\cite{yang2025qwen3} for text conditioning.

\vspace{0.3cm}
\begin{wrapfigure}[14]{r}[0pt]{0.40\textwidth}
\vspace{-0.4cm}
\begin{adjustbox}{width=1\linewidth,center}
\begin{tabular}{c}
    \includegraphics[width=1\textwidth]{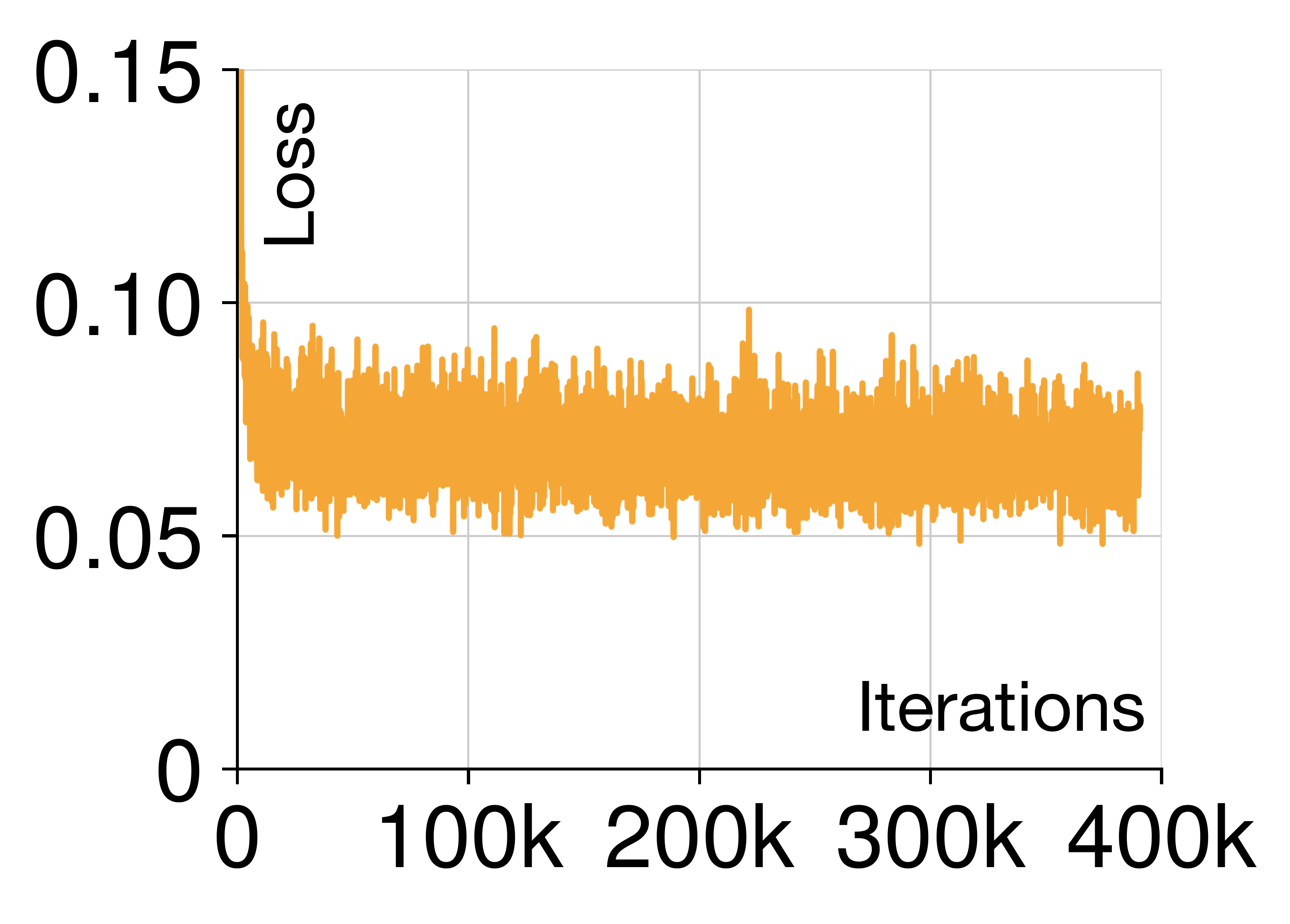} \\
\end{tabular}
\end{adjustbox}
\vspace{-0.4cm}
\caption{
\textbf{Pretraining loss for the JiT-T2I reference baseline \cite{ma2026pixelgen} on GPIC-Full.} 
We show training loss vs. iterations. The model is trained for one epoch on GPIC-Full (100M text-image pairs).
}
\vspace{0.2cm}
\label{fig_main:jit_loss_results}
\end{wrapfigure}
\textbf{Experiment Setup.}
We train JiT-T2I on GPIC-Full for one epoch at $256 \times 256$ resolution. The global batch size is 256. We use AdamW with learning rate $10^{-4}$, betas 0.9 and 0.95, and no weight decay. We use a constant learning-rate schedule with 0.1\% warmup. During training, images are randomly cropped by sampling a crop scale between 0.8 and 1.0 of the original image, followed by a random square crop resized to $256 \times 256$. The maximum text length is 300 tokens. Training took approximately 40 hours on a single 8$\times$H100 node\specialfootnote{~Due to streaming and prefetching errors during distributed training, a small number of samples were repeated.}. For evaluation, we follow the GPIC benchmarking protocol and generate images for the released 50K test captions. We sample with Euler sampling using 50 steps and evaluate classifier-free guidance (CFG) scales 1.75, 4.0, and 6.25.

\begin{figure}[!hb]
    \centering
    \vspace{0.6cm}
    \includegraphics[width=\linewidth]{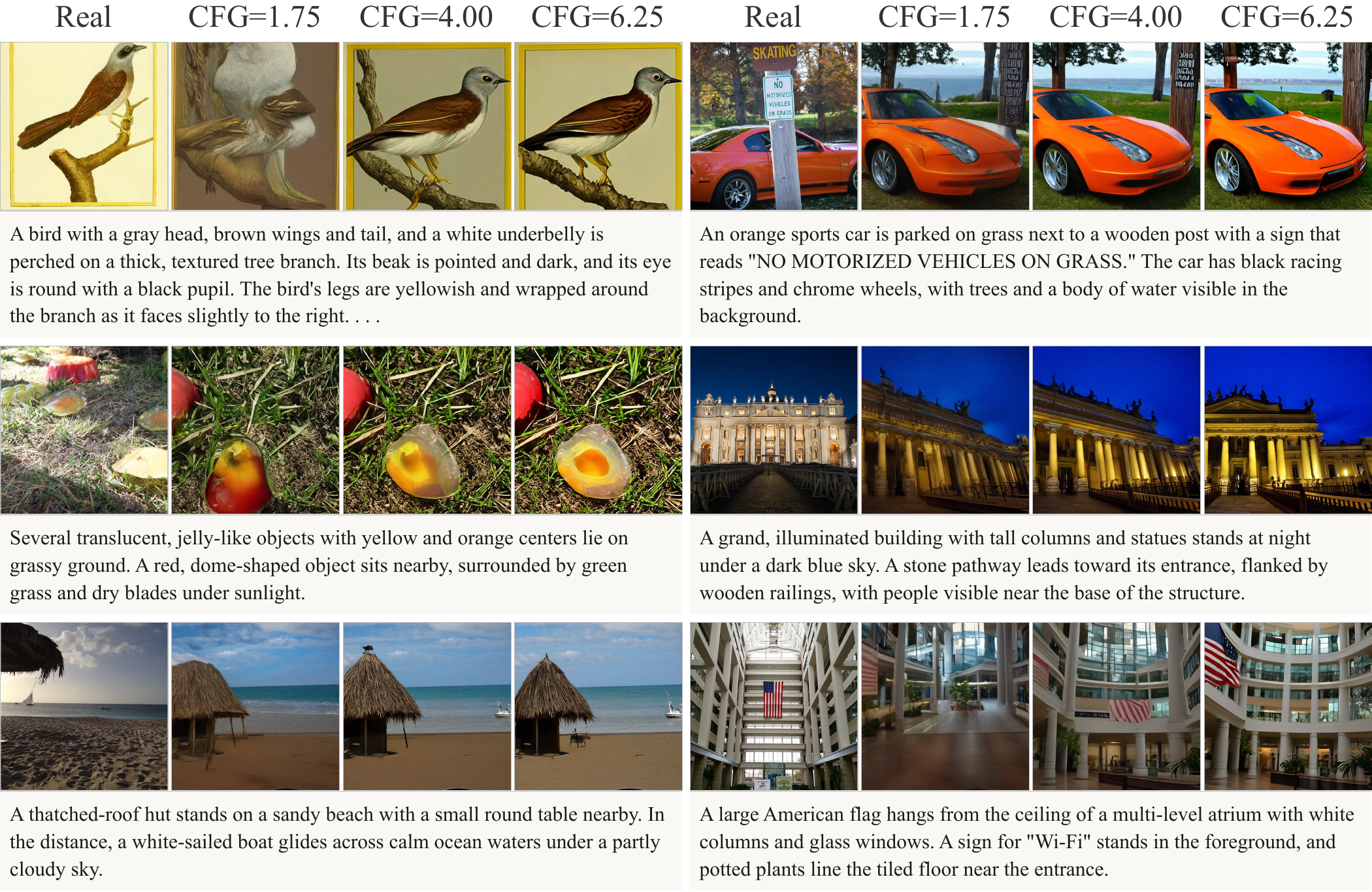}
   \caption{
   \textbf{JiT-T2I samples after training on GPIC-Full for one epoch.}
    We show generated images for prompts in the held-out Test-50K subset.
    Each group contains a real test image, the corresponding text prompt, and JiT-T2I generations sampled with classifier-free guidance scales \(\mathrm{CFG}=1.75, 4.00,\) and \(6.25\), respectively.
    The examples span diverse object-centric and scene-level prompts, including animals, vehicles, natural scenes, architecture, and indoor environments.
    Quantitative results for each CFG scale are reported in Table~\ref{tab_main:jit_pretraining_results}.
    }
    \label{fig:jit-samples}    
\end{figure}

\vspace{0.3cm}
\textbf{Results.}
We report quantitative results in Table~\ref{tab_main:jit_pretraining_results}, the pretraining loss curve in Figure~\ref{fig_main:jit_loss_results}, and qualitative samples in Figure~\ref{fig:jit-samples}.
The baseline achieves its best FD of \(76.25\) at CFG \(6.25\).
Increasing CFG improves FD, recall, and coverage in this baseline.
The best CFG value under FD-DINOv2 on GPIC is higher than values commonly used for class-conditional ImageNet-1K evaluation with FID.
We release the model as a reference baseline for future comparisons on GPIC.

\begin{table}[h]
\centering

\begin{adjustbox}{width=0.7\linewidth}
\Large

\begin{tabular}{crcccc}
\textbf{CFG} &\textbf{FD} $\downarrow$ &\textbf{Precision} $\uparrow$ &\textbf{Recall} $\uparrow$ &\textbf{Density} $\uparrow$ &\textbf{Coverage} $\uparrow$ \\
\toprule
1.75 &204.01 &0.917 &0.530 &1.034 &0.806 \\
\midrule
4.00 &87.80 &0.933 &0.765 &1.012 &0.906 \\
\midrule
6.25 &76.25 &0.942 &0.792 &1.014 &0.908 \\
\bottomrule
\end{tabular}
\end{adjustbox}
\vspace{0.2cm}
\caption{
\textbf{JiT-T2I baseline results on GPIC-Full after training for one epoch.} 
We report FD, Precision, Recall, Density, and Coverage for three classifier-free guidance scales. We use 50-step Euler sampling for all generations. All metrics are computed against the 1M GPIC test set.
}
\vspace{-2mm}
\label{tab_main:jit_pretraining_results}
\end{table}

\section{Conclusion}
\label{sec:conclusion}
GPIC (Giant Permissive Image Corpus) is a large, permissive benchmark dataset for visual generative modeling. In this paper, we described the design choices needed to make GPIC permissive, stable, large, and accessible, including its construction pipeline, release format, evaluation protocol, compliance guidelines, and reference baseline. As visual generative models continue to evolve, benchmark datasets and metrics must evolve with them. Beyond text-to-image generation, GPIC provides a large-scale, high-quality image-text resource for broader multimodal research. We hope GPIC supports open, accessible, and reproducible research on large-scale visual generative modeling.
GPIC is available at \href{https://huggingface.co/datasets/stanford-vision-lab/gpic}{Hugging Face}, and the evaluation toolkit and PyTorch code are available at \href{https://gpic.stanford.edu}{gpic.stanford.edu}.

\vspace{0.25cm}

\textbf{Broader Impact and Limitations.}
GPIC is a fully permissive 100M-image dataset for visual generative modeling, supporting transparent and legally verifiable benchmarking and model training. At the same time, GPIC carries societal risks shared with prior large-scale image corpora~\cite{yfcc100m, laion, datacomp}, including potential misuse for harmful generation, memorization of training content, and amplification of source-platform biases. We take several steps to mitigate these risks. Every image in GPIC has a clear legal basis for redistribution and use, and license names, license URLs, and attribution strings are retained as metadata for every sample. Captions are generated by Qwen3-VL-4B~\cite{bai2025qwen3vl} rather than scraped from alt text, avoiding direct release of source text that may contain toxic or personally identifying language. We also release GPIC as frozen tar shards rather than a URL index, eliminating silent dataset drift, exposure to URL-level data poisoning, and the need to re-scrape source images outside our filtering pipeline. Finally, despite our deduplication efforts, some near-duplicates may remain in GPIC, although their prevalence is estimated to be small.
\section*{Acknowledgments}
We thank Radical Numerics and World Labs for providing compute for this project.
We thank Willie Neiswanger, Yue Zhao, Armin W. Thomas, Garyk Brixi, Manling Li, Tristan Thrush, Bailey Trang, and Aryaman Arora for their feedback on the manuscript.
We thank Agrim Gupta for valuable discussions.
We thank members of the Stanford Vision Lab and the CogAI group for their feedback.

\clearpage
\bibliographystyle{unsrtnat}
\bibliography{references}
\clearpage
\section*{Appendix}
\appendix
\renewcommand{\thefigure}{\thesection.\arabic{figure}}
\renewcommand{\thetable}{\thesection.\arabic{table}}

\numberwithin{figure}{section}
\numberwithin{table}{section}

\etocdepthtag.toc{appendix}
\etocsettocstyle{\textbf{Contents}}{}
\tableofcontents

\hiddensection{Additional Image-Text examples from GPIC}

\begin{figure}[!t]
  \thispagestyle{empty}
  \centering
\includegraphics[width=\textwidth]{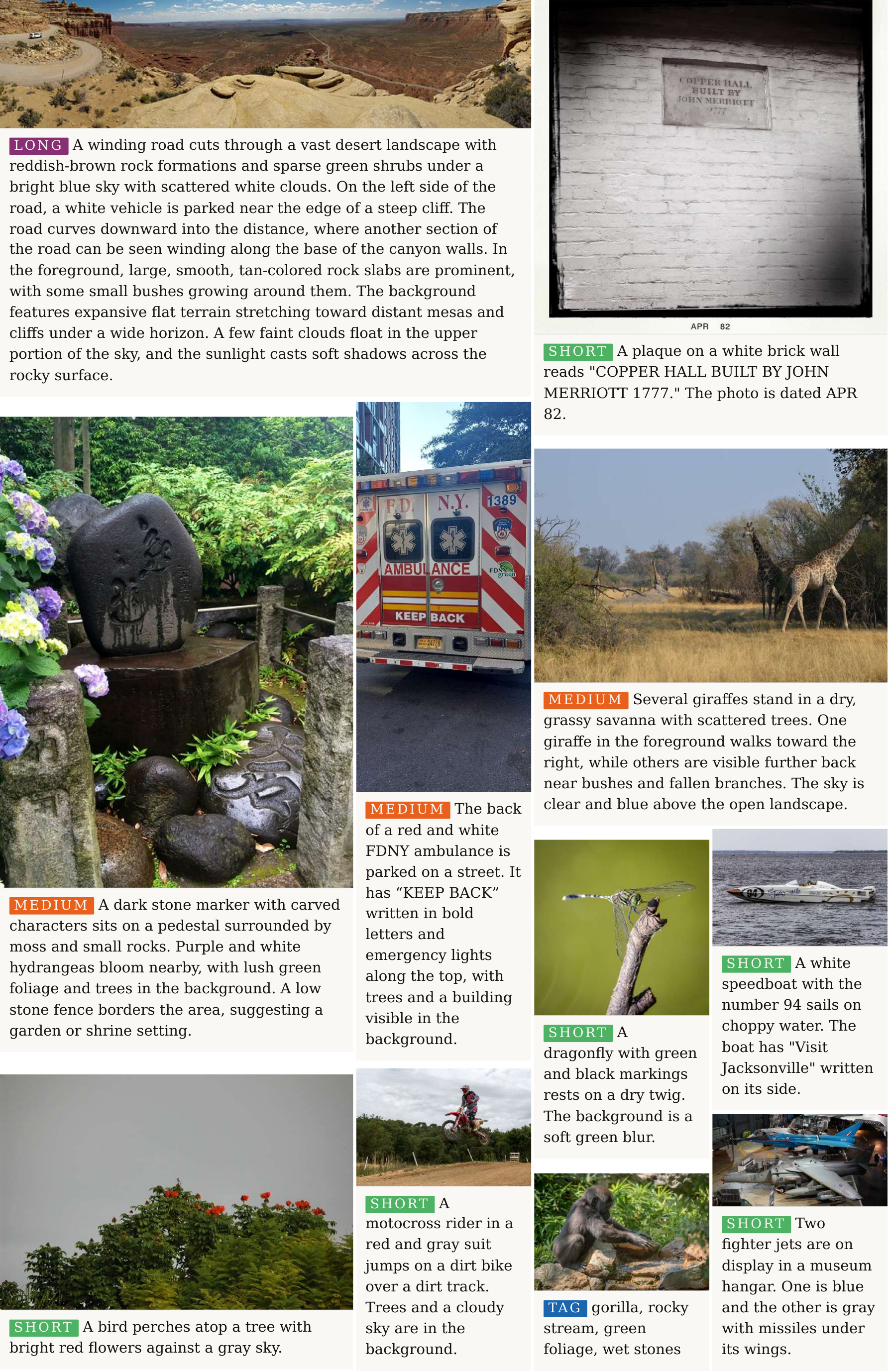}
  \caption{Additional example image-caption pairs from GPIC.}
  \label{fig:gpic_samples}
\end{figure}
\FloatBarrier

\section{Evaluation}
\label{sec_supp:evals}

\subsection{Construction of Imagenet-256 and GPIC-256}

We adopt the following protocol:

\begin{enumerate}
    \item Center crop along the longer edge to form a square image.
    \item Bicubic downsampling to 256 $\times$ 256 from the Pillow library \cite{clark2015pillow}.
\end{enumerate}

We note that popular Python image libraries use different bicubic interpolation kernels. Our choice of Pillow is consistent with prior work \cite{stein2023exposing, parmar2021cleanfid}.

\subsection{Additional Oracle Reference Metrics}

\begin{table}[h]
\centering
\begin{tabular}{lccccc}
\toprule
GPIC Subset & FD $\downarrow$ & Precision $\uparrow$ & Recall $\uparrow$ & Density $\uparrow$ & Coverage $\uparrow$ \\
\midrule
Full & 0.07  & 0.757 & 0.762 & 0.973 & 0.966 \\
Lite & 0.07 & 0.762 & 0.768 & 1.041 & 0.973 \\
Nano & 0.11 & 0.756 & 0.769 & 1.000 & 0.973 \\
Val  & 0.21 & 0.760 & 0.758 & 1.007 & 0.971 \\
Test-50K & 0.68 & 0.7688 & 0.763 & 0.979 & 0.964 \\
\bottomrule
\end{tabular}%
\vspace{2mm}
\caption{Generative quality metrics over Inception-v3 representations across GPIC subsets against GPIC-Test-1M. We omit MMD as each subset scores $\approx 0$.}
\label{tab:gpic-subsets-metrics-inception}
\end{table}

\begin{table}[h]
\centering
\begin{tabular}{lcccc}
\toprule
& \multicolumn{2}{c}{DINOv2} & \multicolumn{2}{c}{Inception-v3} \\
\cmidrule(lr){2-3} \cmidrule(lr){4-5}
GPIC Subset & $\mathrm{FD}_{\mu}$ $\downarrow$ & $\mathrm{FD}_{\Sigma}$ $\downarrow$ & $\mathrm{FD}_{\mu}$ $\downarrow$ & $\mathrm{FD}_{\Sigma}$ $\downarrow$ \\
\midrule
Full & 0.051 & 1.140 & 0.005 & 0.065  \\
Lite & 0.052 & 1.197 & 0.005 & 0.069  \\
Nano & 0.053 & 1.547 & 0.005 & 0.100  \\
Val & 0.012 & 2.353 & 0.001 & 0.208  \\
Test-50K & 0.040 & 7.404 & 0.004 & 0.678  \\
\bottomrule
\end{tabular}%
\vspace{2mm}
\caption{$\mathrm{FD}_{\mu}$ and $\mathrm{FD}_{\Sigma}$ across GPIC subsets against GPIC-Test-1M.}
\label{tab:fd-decomp}
\end{table}
\FloatBarrier

\subsection{Effect of DINOv2 Backbone Size and Register Tokens on FD}

\begin{figure}
    \centering
    \includegraphics[width=\linewidth]{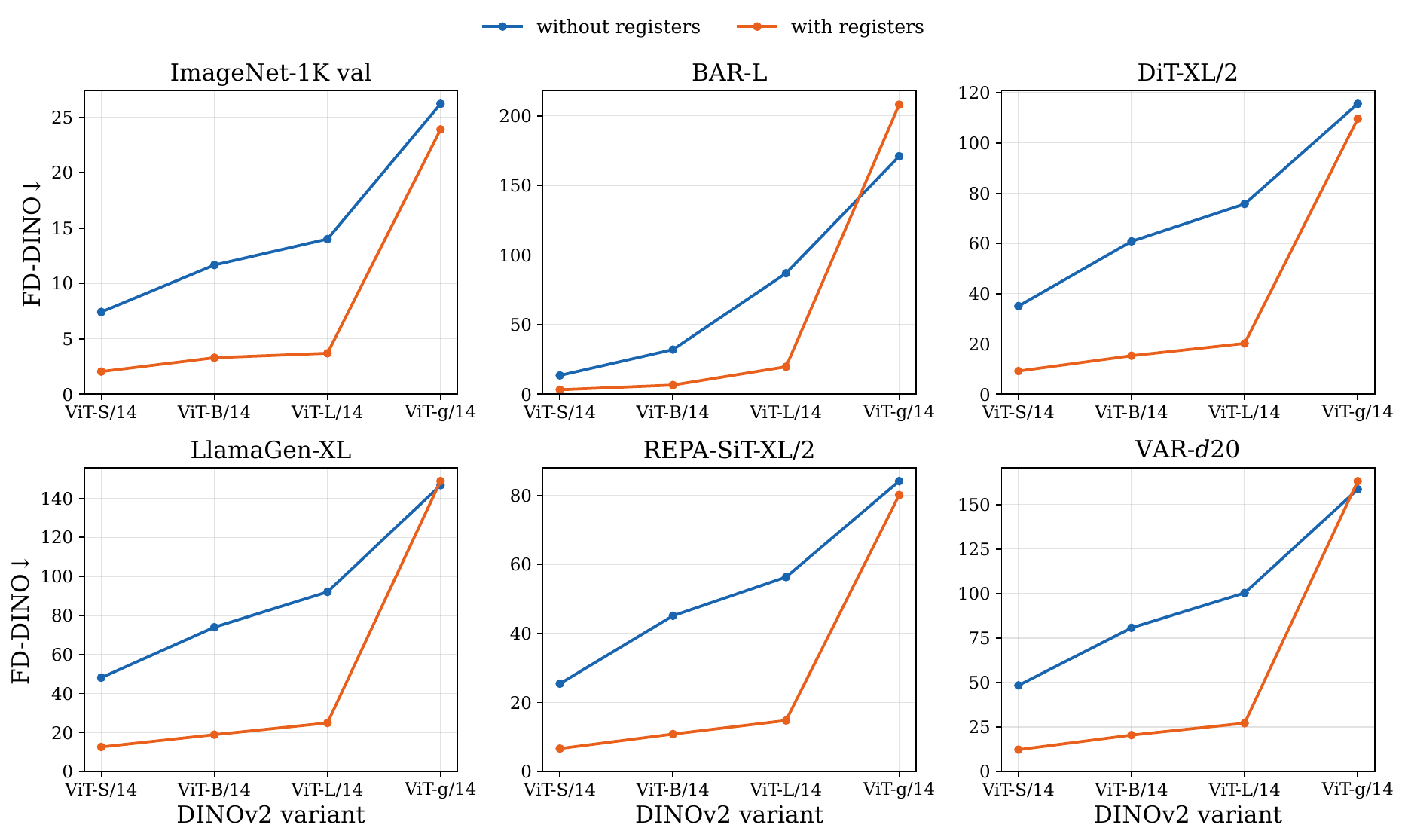}
    \caption{FD-DINOv2 across DINOv2 model sizes and variants with and without registers.}
    \label{fig:fd-dino-variants}
\end{figure}

\begin{figure}
    \centering
    \includegraphics[width=\linewidth]{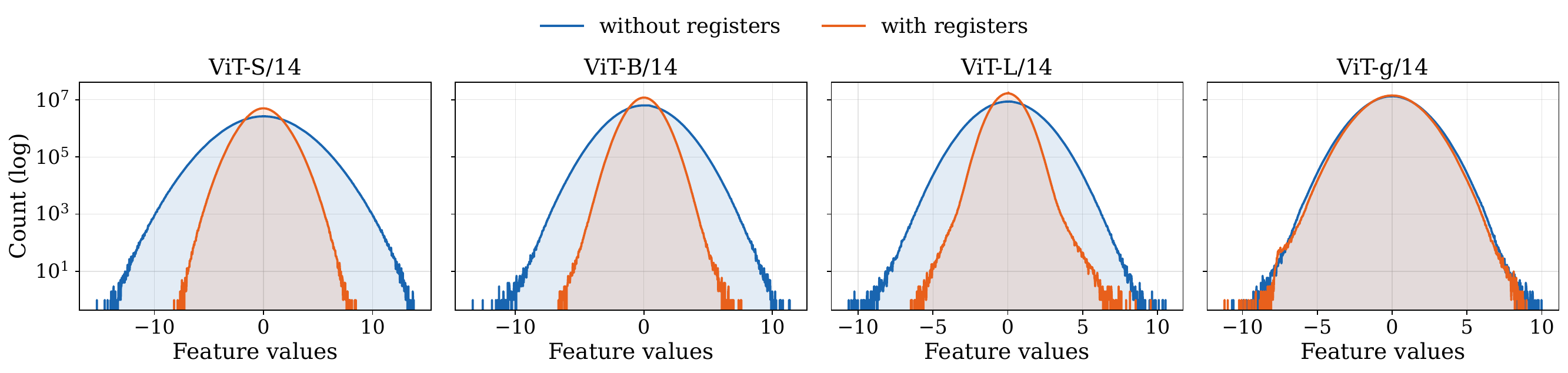}
    \caption{Distribution of DINOv2 feature values on ImageNet-1K train images.}
    \label{fig:register-hist}
\end{figure}

A practical question when using FD-DINOv2 is how to interpret its numerical scale.
Unlike pixel-space distances, FD-DINOv2 depends on feature values produced by a learned neural network.
In particular, these feature values can change with the DINOv2 backbone size and whether the model uses registers.
We therefore ablate variants with and without registers across four DINOv2 backbone sizes: ViT-S/14, ViT-B/14, ViT-L/14, and ViT-g/14.

Registers in vision transformers were introduced to reduce high-norm artifacts in DINOv2 feature maps.
Since FD-DINOv2 is computed directly in DINOv2 feature space, changes to the feature distribution can affect both the absolute metric value and comparisons between generative models.

Results are shown in Figure~\ref{fig:fd-dino-variants}.
For the small, base, and large backbones, variants with registers consistently produce lower FD-DINOv2 scores than the corresponding variants without registers.
At the giant scale, the variants with and without registers are much closer.

The feature-value histograms in Figure~\ref{fig:register-hist} help explain this pattern.
For the small, base, and large backbones, variants with registers produce feature values over a much smaller range than the corresponding variants without registers.
Since Fr\'echet distance depends on both feature means and covariances, changes in the range and variance of feature values directly affect the scale of FD-DINOv2.
In contrast, the ViT-g/14 distributions with and without registers nearly overlap, matching the smaller FD-DINOv2 difference at the giant scale.

Despite these differences in absolute metric scale, the variants remain broadly consistent as evaluators.
Across the eight DINOv2 configurations, the mean pairwise Pearson correlation over the five models and ImageNet-1K validation set is \(0.847\), and Kendall's coefficient of concordance over the five models is \(0.795\).
These values indicate strong agreement in the relative ordering of models.
Following prior generative model evaluation work that uses DINOv2 ViT-L/14 features~\citep{stein2023exposing}, we use the non-register ViT-L/14 backbone as the default FD-DINOv2 configuration.
We leave a dedicated human-alignment study of variants with registers to future work.
\clearpage

\section{Image Filtering}

\begin{figure}[h]
    \centering
    \includegraphics[width=0.75\linewidth]{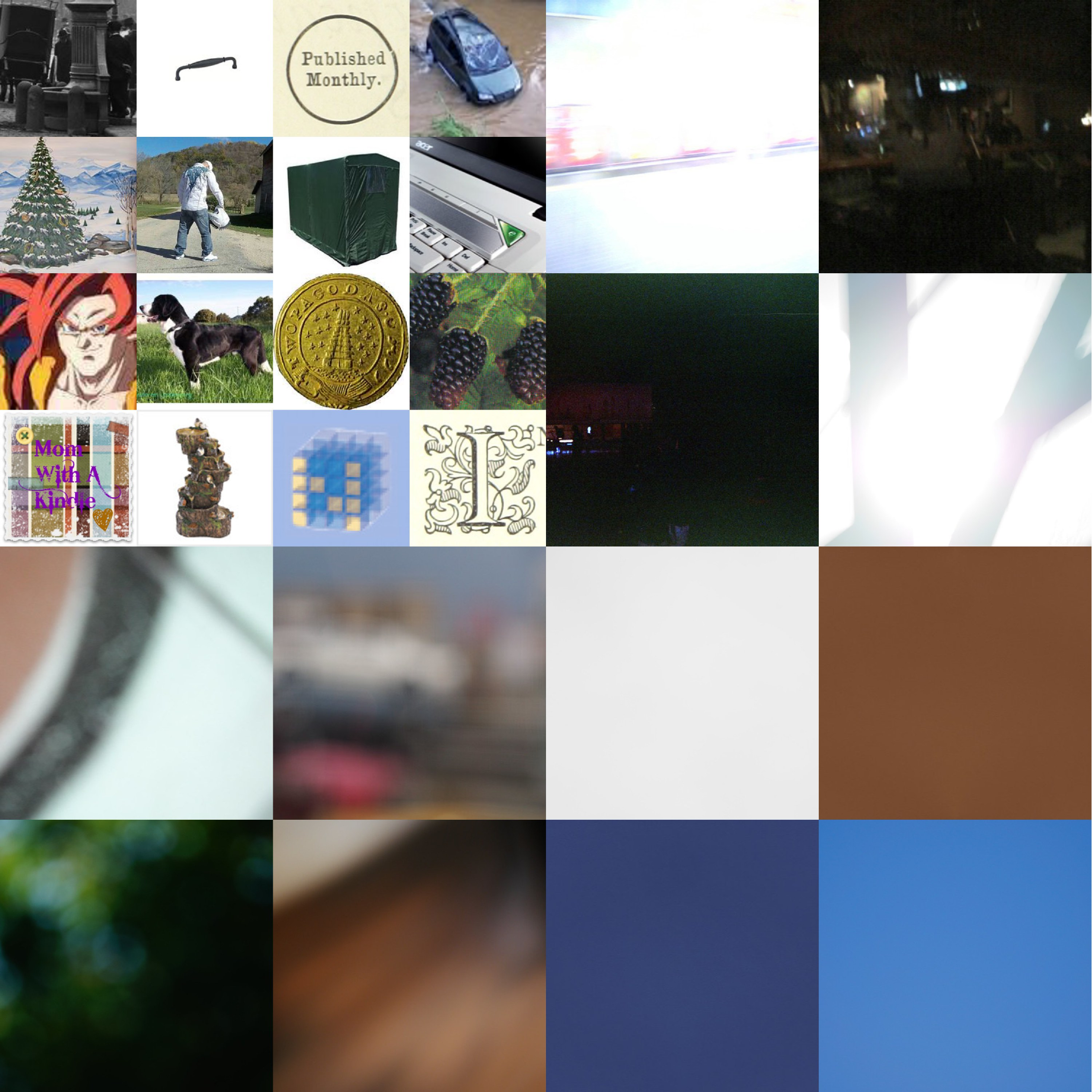}
    \caption{Additional low resolution and poor visual quality image examples.}
    \label{fig:filtered-supp}
\end{figure}
\FloatBarrier

\section{Deduplication}
\begin{figure}[h]
    \centering
    \includegraphics[width=\linewidth]{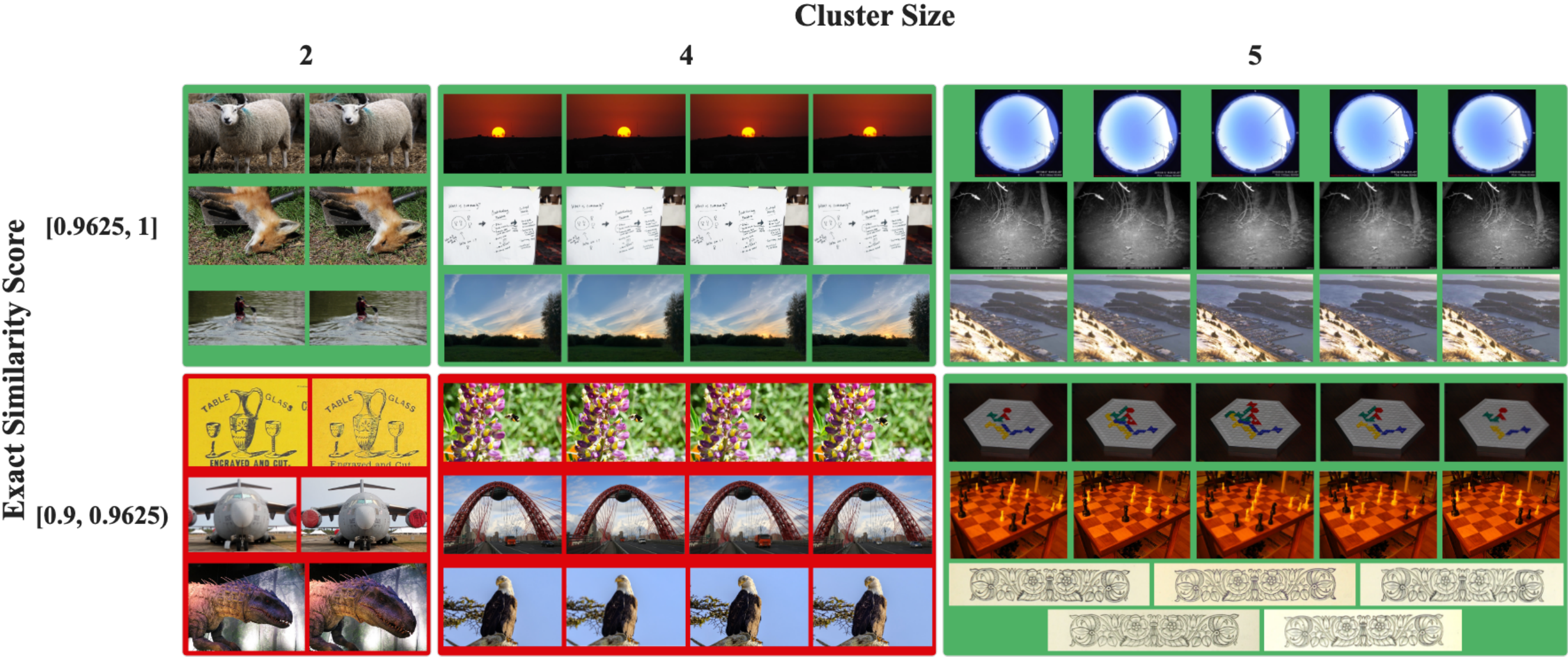}
    \caption{Qualitative examples of deduplication over similarity score tiers and cluster sizes. All clusters with exact similarity score $\geq 0.9625$ are removed, and only clusters of size $\geq 5$ are removed for similarity scores in $[0.9, 0.9625)$.}
    \label{fig:dedup-examples}
\end{figure}
\FloatBarrier

\section{Microbenchmark}
\label{sec_supp:microbenchmark}
We sampled 1,520 images from the initial source pooling stage of our dataset construction for our microbenchmark to evaluate the captioning quality of Qwen3-VL-Instruct models. From the initial VLM captions, human annotators examined and relabelled the captions, fixing any errors and hallucinations. Common errors included counting and spatial relations. Examples of VLM-labeled and human-labeled caption pairs are shown in Fig.~\ref{fig:microbenchmark-samples}. 
The final human-labeled captions were used as ground truth labels for our LLM-as-a-judge pipeline to evaluate the captioning quality of the Qwen3-VL-Instruct models.

\begin{figure}[h]
    \includegraphics[width=\linewidth]{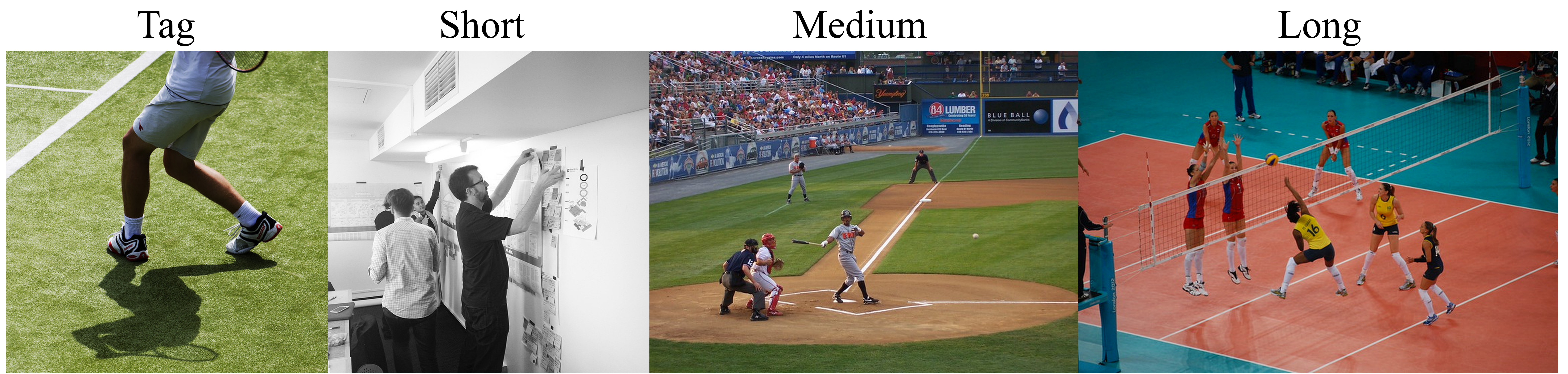}
{\small
\textbf{Tag:}

\textcolor{red}{Model:} tennis court, athletic shoes, grass, shadow, \underline{player}.

\textcolor{blue}{Human:} tennis court, athletic shoes, grass, shadow, \underline{lower body}.

\vspace{1em}
\textbf{Short:}

\textcolor{red}{Model:} A group of people \underline{looks at papers on a wall in an office}.

\textcolor{blue}{Human:} A group of people \underline{pins papers and sticky notes to an office wall}.

\vspace{1em}
\textbf{Medium:}

\textcolor{red}{Model:} A baseball batter in a gray uniform stands at home plate after swinging, with the ball traveling across the field to the right. A catcher in red gear and an umpire crouch behind him, while another player stands near the \underline{first-base} line and fans fill the stadium seats in the background.

\textcolor{blue}{Human:} A baseball batter in a gray uniform stands at home plate after swinging, with the ball traveling across the field to the right. A catcher in red gear and an umpire crouch behind him, while another player stands near the \underline{third-base} line and fans fill the stadium seats in the background.

\vspace{1em}
\textbf{Long:}

\textcolor{red}{Model:} \underline{Five} female volleyball players are positioned on a red and teal court, engaged in a match. \underline{Three} players in red and blue uniforms are on the left side of the net, with two raising their arms to block a yellow volleyball that is mid-air near the net. One player in a yellow jersey with the number 16 is in a low crouch, facing the net, and another player in a yellow jersey is standing to the right\underline{, near the sideline with arms extended}. A fifth player in a black and yellow uniform is positioned further right, standing with her feet apart and looking toward the ball. The net is stretched across the center of the court, and a blue vertical post with \underline{the words ``Olympic Games''} visible is on the right side of the net. In the background, spectators are seated on benches along the sidelines, and a blue umpire's chair is visible on the far left. The court surface is marked with white lines, and the net has a white top band with vertical markings.

\textcolor{blue}{Human:} \underline{Seven} female volleyball players are positioned on a red and teal court, engaged in a match. \underline{Four} players in red and blue uniforms are on the left side of the net, with two raising their arms to block a yellow volleyball that is mid-air near the net. One player in a yellow jersey with the number 16 is in a low crouch, facing the net, and another player in a yellow jersey is standing to the right. A fifth player in a black and yellow uniform is positioned further right, standing with her feet apart and looking toward the ball. The net is stretched across the center of the court, and a blue vertical post with \underline{``London 2012''} visible is on the right side of the net. In the background, spectators are seated on benches along the sidelines, and a blue umpire's chair is visible on the far left. The court surface is marked with white lines, and the net has a white top band with vertical markings.
}
\vspace{1em}
\caption{Full caption comparison between VLM-generated and human-labeled annotations. Red highlights VLM captioning model outputs and blue highlights human annotations. Underlined text indicates the specific differences in counting, spatial relations, and fine-grained visual details.}
\end{figure}
\label{fig:microbenchmark-samples}
\FloatBarrier

\section{Prompts}

\begin{figure}
    \centering
    \includegraphics[width=\linewidth]{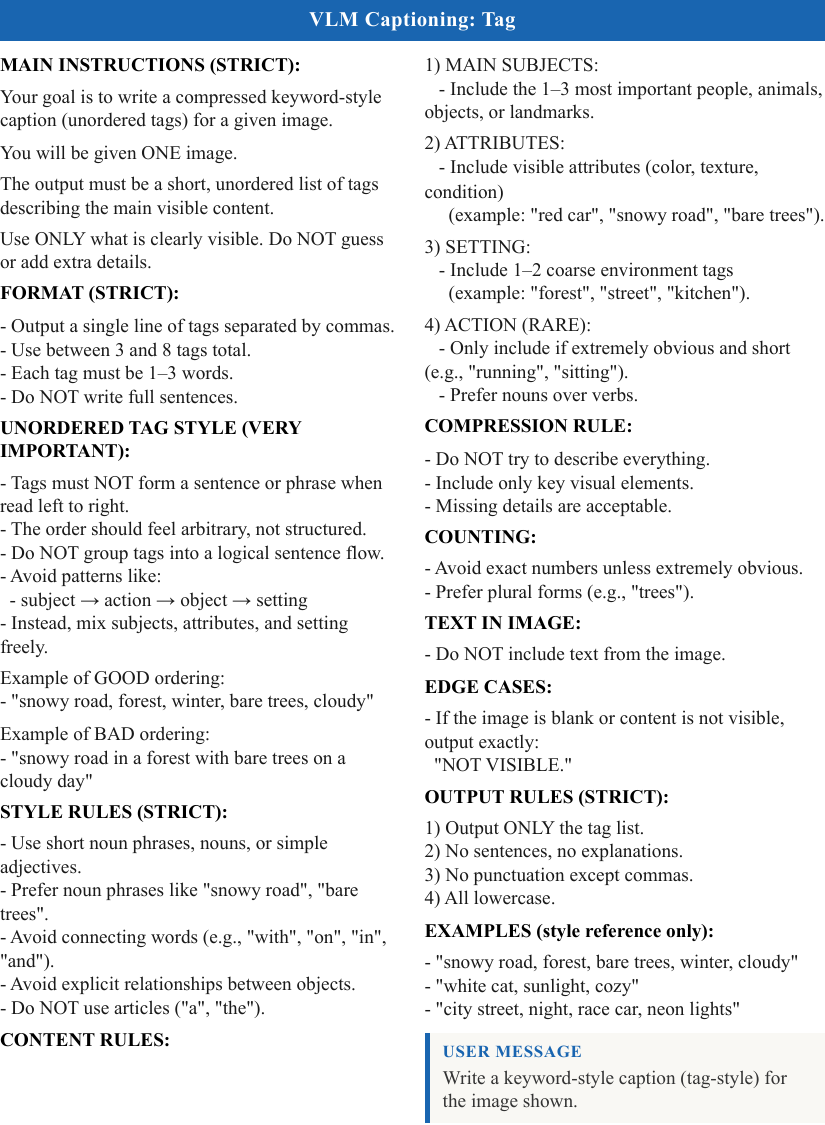}
    
    \caption{Prompt used to generate tag-styled captions for GPIC images.}
    \label{fig:prompt-caption-tag}
\end{figure}

\begin{figure}
    \centering
    \includegraphics[width=\linewidth]{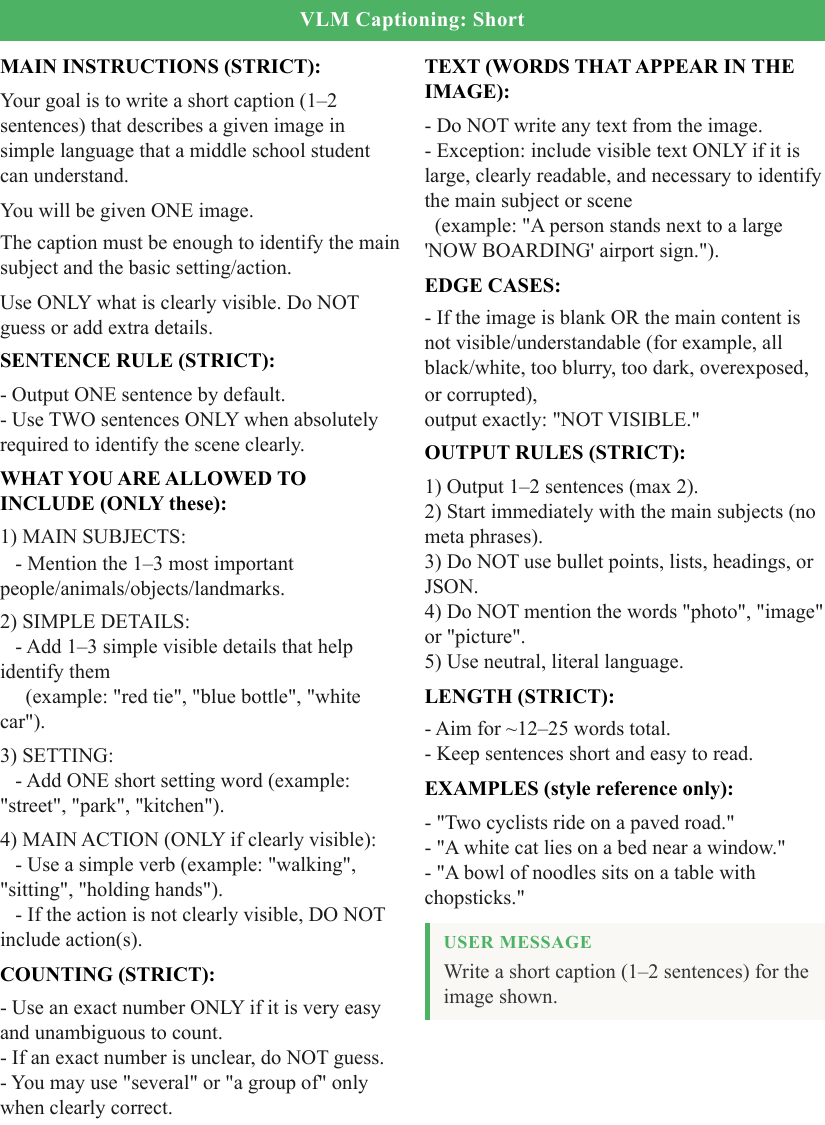}
    \caption{Prompt used to generate short-length captions for GPIC images.}
    \label{fig:prompt-caption-short}
\end{figure}

\begin{figure}
    \centering
    \includegraphics[width=\linewidth]{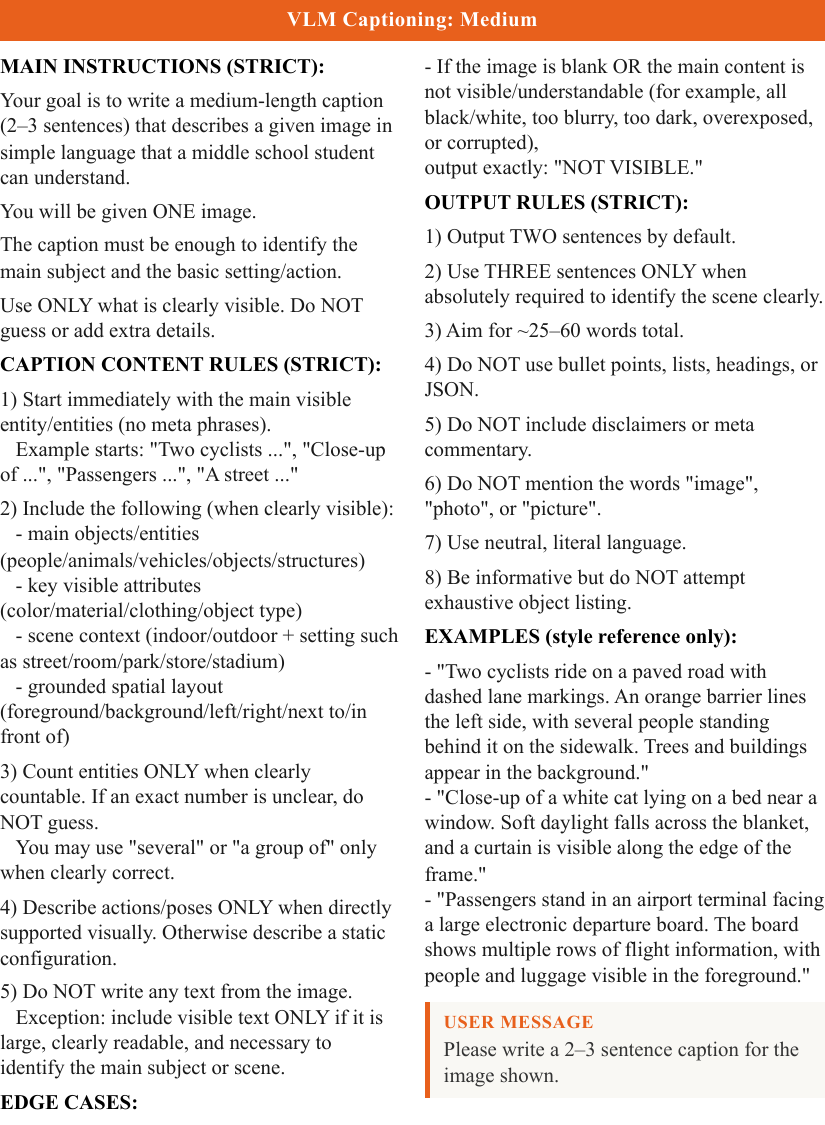}
    
    \caption{Prompt used to generate medium-length captions for GPIC images.}
    \label{fig:prompt-caption-medium}
\end{figure}

\begin{figure}
    \centering
    \includegraphics[width=\linewidth]{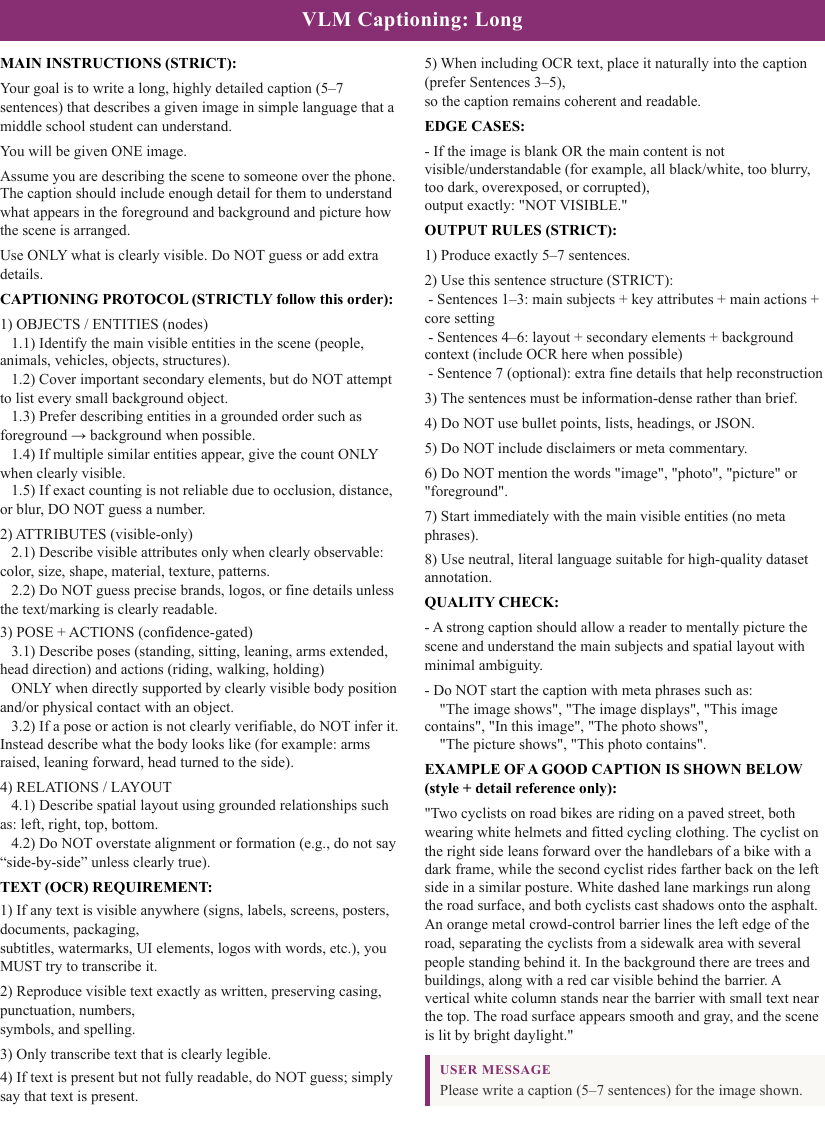}
    
    \caption{Prompt used to generate long-sized captions for GPIC images.}
    \label{fig:prompt-caption-long}
\end{figure}

\begin{figure}
    \centering
    \includegraphics[width=\linewidth]{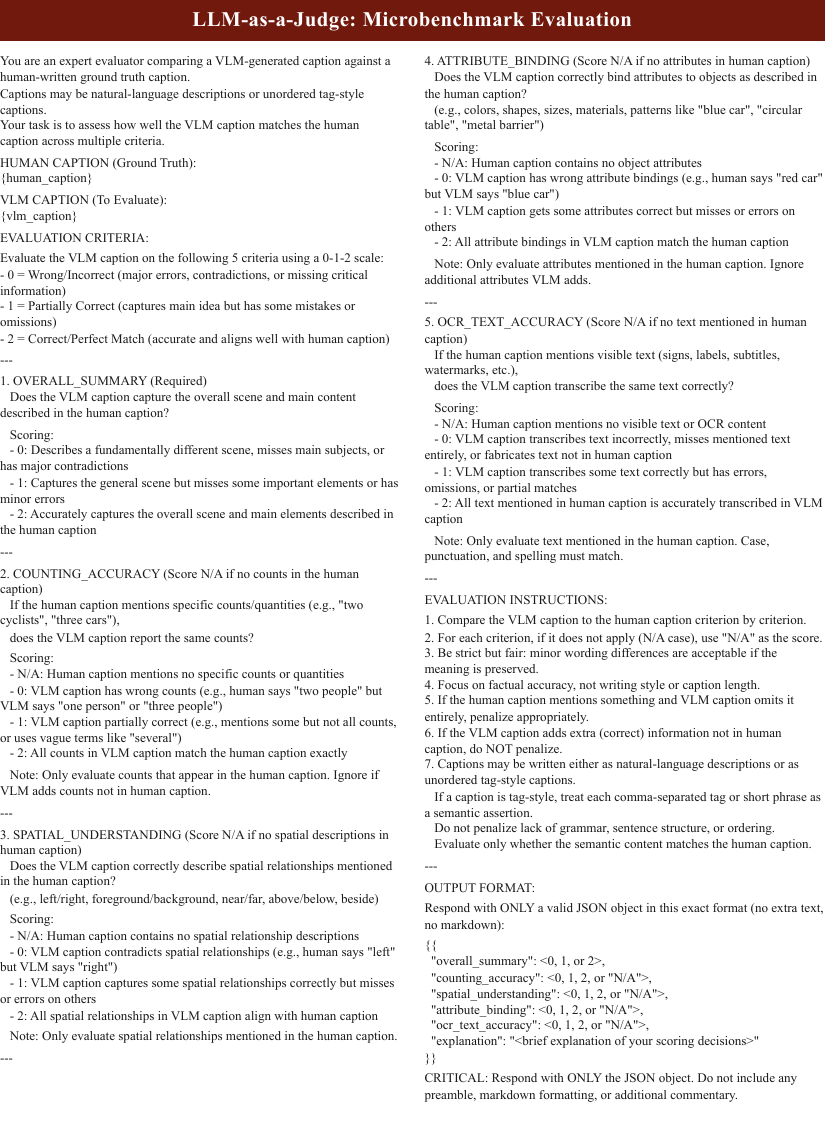}
    
    \caption{Prompt used to evaluate VLM-generated captions against ground-truth captions in our microbenchmark.}
    \label{fig:prompt-llm-judge}
\end{figure}

We include the exact prompts used for the following tasks:
\begin{itemize}
    \item Tag-style GPIC image captioning (Fig.~\ref{fig:prompt-caption-tag}).
    \item Short-form GPIC image captioning (Fig.~\ref{fig:prompt-caption-short}).
    \item Medium-length GPIC image captioning (Fig.~\ref{fig:prompt-caption-medium}).
    \item Long-form GPIC image captioning (Fig.~\ref{fig:prompt-caption-long}).
    \item Evaluating VLM-generated captions against ground-truth captions in our microbenchmark (Fig.~\ref{fig:prompt-llm-judge}).
\end{itemize}

\FloatBarrier


\end{document}